\documentclass[10pt,journal]{IEEEtran} 
\IEEEoverridecommandlockouts

\usepackage{array}
\newcolumntype{L}[1]{>{\raggedright\arraybackslash}p{#1}}
\usepackage{amsmath,amssymb,amsfonts}
\usepackage{mathtools}
\usepackage{amsthm}

\usepackage{graphicx}
\usepackage[dvipsnames]{xcolor}
\usepackage{centernot}
\def\BibTeX{{\rm B\kern-.05em{\sc i\kern-.025em b}\kern-.08em T\kern-.1667em\lower.7ex\hbox{E}\kern-.125emX}}  
\usepackage{array}
\usepackage{textcomp}
\usepackage{stfloats}
\usepackage{ragged2e}
\usepackage{algpseudocode}
\usepackage{url}
\usepackage{longtable}
\usepackage{physics}
\usepackage{placeins}
\usepackage{tabularx}
\usepackage{verbatim}
\usepackage{enumitem}
\usepackage{stfloats}
\usepackage[bottom]{footmisc}
\usepackage{booktabs}
\usepackage{multirow}
\usepackage[nocompress]{cite}
\usepackage[colorlinks=true, citecolor=blue, linkcolor=black, urlcolor=blue]{hyperref}
\newcolumntype{L}[1]{>{\RaggedRight\arraybackslash}p{#1}}
\newcolumntype{Y}{>{\RaggedRight\arraybackslash}X}
\makeatletter
\renewcommand{\@cite}[2]{\textcolor{blue}{[{#1\if@tempswa , #2\fi}]}}
\makeatother

\usepackage{siunitx} 
\usepackage[bottom]{footmisc}
\usepackage{mathtools}
\usepackage{subcaption}
\usepackage{algorithmicx}
\usepackage{algpseudocode}
\algrenewcommand{\algorithmiccomment}[1]{\hfill// #1}
\algnewcommand{\algorithmicor}{\textbf{ or }}
\algnewcommand{\OR}{\algorithmicor}
\usepackage{balance}
\usepackage{algorithm}
\usepackage{makecell}

\title{RBLA+: Efficient Rank-Specific LoRA Adapter Extraction for Federated Fine-Tuning with Heterogeneous Clients}

\author{Shuaijun~Chen,~Omid~Tavallaie,~Xin~Chen,~Wentao~Gao,~Nguyen~H.~Tran,~and~Albert~Y.~Zomaya%
\IEEEcompsocitemizethanks{%
\IEEEcompsocthanksitem Shuaijun Chen, Wentao Gao, Nguyen H. Tran, and Albert Y. Zomaya are with the School of Computer Science, The University of Sydney, Sydney, NSW, Australia.
E-mail: \{shuaijun.chen, wentao.gao, nguyen.tran, albert.zomaya\}@sydney.edu.au.
\IEEEcompsocthanksitem Omid~Tavallaie is with the Department of Computer Science, The University of Western Australia, Perth, Australia, and is also a Visiting Academic with the Department of Engineering Science, University of Oxford, United Kingdom. Email: omid.tavallaie@uwa.edu.au.
E-mail: omid.tavallaie@uwa.edu.au.
\IEEEcompsocthanksitem Xin Chen is with the School of Mathematics and Statistics, Central South University, China.
E-mail: chenxin@csu.edu.cn.}%
}

\begin{document}

\IEEEtitleabstractindextext{%
\begin{abstract}
Federated fine-tuning with Low-Rank Adaptation (LoRA) reduces communication and computational overhead by training and exchanging compact trainable LoRA factors. Clients may use different LoRA ranks to meet adapter communication budgets and accommodate different local adaptation needs. The server must process these differences during aggregation and construct adapters at each client's rank for local training. We distinguish pre-aggregation alignment, which handles factor dimensions and scaling conventions before aggregation and rank-specific LoRA adapter extraction, which constructs client-sized pairs from the aggregated global model. Two approaches to rank-specific extraction are direct rank truncation and Singular Value Decomposition (SVD) of the dense update. Direct prefix truncation keeps the leading columns of one aggregated factor and the corresponding rows of the other. It is cheap but increases reconstruction error. The dense SVD approach forms the product of aggregated LoRA factors. It then retains the leading singular components to construct adapters at the target rank. This gives the optimal rank-constrained approximation in Frobenius norm but incurs significant overhead. To address this trade-off, we propose RBLA+ as an extension of RBLA. It uses compact QR/core-SVD decomposition to minimize Frobenius error without forming the dense update, and aligns client and server scaling conventions during upload and broadcast. Evaluation results revealed that compared to state-of-the-art methods, RBLA+ can improve accuracy by up to 3.30\%, and by up to 40.79 percentage points over RBLA on CIFAR-10. In CPU benchmarks with synthetic adapters, it extracts adapters faster than the dense partial-SVD baseline while preserving each client's prescribed adapter size and per-round communication overhead.
\end{abstract}

}

\maketitle
\IEEEdisplaynontitleabstractindextext

\section{Introduction}
\label{sec:introduction}

Federated Learning (FL) allows clients to train a shared model without collecting their training data at a central server~\cite{mcmahan2017fedavg}. Clients perform local training, upload model updates, and receive an aggregated model for the next round. Frameworks such as NVIDIA FLARE~\cite{roth2021flare} and Flower~\cite{flowerFramework} support this collaborative workflow. For resource-constrained mobile and edge clients, however, the cost of participating can be substantial. Fine-tuning a large model requires substantial memory for model weights, gradients, optimizer states, and intermediate activations. Repeated full-model exchanges between clients and the server also consume communication capacity. Low-Rank Adaptation (LoRA) reduces these demands by freezing pretrained weights and training and exchanging compact adapters. Clients can choose different LoRA ranks, which control adapter sizes, to accommodate their resource budgets. This paper studies how the server coordinates these adapters of different sizes. The goal is to reconstruct updates within each client's LoRA rank budget while keeping server processing costs low.

\begin{figure}[t]
  \centering
    \includegraphics[width=0.9\linewidth]{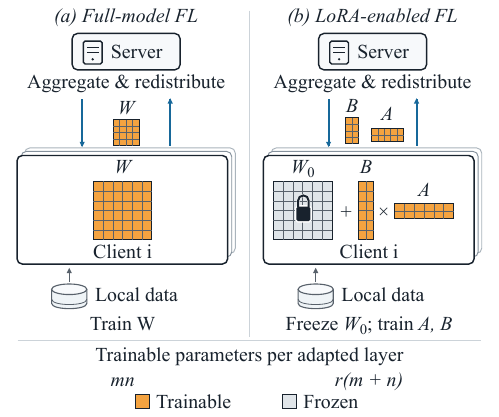}
  \caption{Full-model FL (left) and LoRA-enabled FL (right) after initial model distribution. Both keep training data local. Full-model FL trains and exchanges all model parameters. LoRA-enabled FL freezes the pretrained weights and trains and exchanges LoRA factors $B$ and $A$. For an $m\times n$ weight matrix, full-weight training and LoRA require $mn$ and $r(m+n)$ trainable parameters, respectively, excluding biases and other trainable modules.}
  \label{fig:fl_lora_overview}\vspace{-4mm}
\end{figure}

LoRA represents each adapted weight update as the scaled product of two small trainable matrices, called LoRA factors $B$ and $A$~\cite{hu2021lora}, whose shared matrix dimension is the LoRA rank. Lower ranks reduce the number of adapter parameters trained and exchanged. Higher ranks can represent a broader set of updates. Existing tools support adapter training and inference. NVIDIA NeMo provides LoRA fine-tuning~\cite{nvidiaNeMo}, while Amazon SageMaker supports inference with multiple adapters that share a base model~\cite{nguyen2024sagemaker, wu2024fedfmsl}. As illustrated in Fig.~\ref{fig:fl_lora_overview}, in LoRA-enabled FL, clients first receive a pretrained dense model and keep its weights frozen during local training. In each communication round, they train their LoRA adapters on local data and upload them to the server. The server aggregates the uploaded updates and broadcasts updated adapters for the next round.

\begin{figure}[t]
  \centering
    \includegraphics[width=0.95\linewidth]{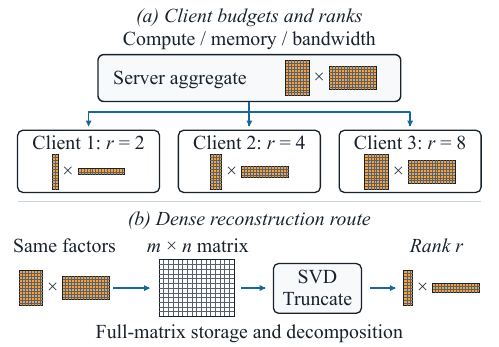}
  \caption{Heterogeneous client ranks and dense adapter extraction. (a) Different ranks allow different adapter sizes. (b) Dense extraction forms the aggregated update and applies SVD to construct LoRA factors at each requested rank. This requires storage and decomposition at the full weight-matrix dimensions.}
  \label{fig:heterogeneous_redistribution}\vspace{-4mm}
\end{figure}

In practice, clients may use different adapter ranks. Their computation, memory, and communication resources can be different, making a fixed large adapter rank not suitable for all clients~\cite{liu2025adaptive}. Assigning a small adapter to every client can limit the adaptation capacity available to clients with higher budgets. Local data can also motivate different ranks. Even among clients with similar resources, differences in data volume and distribution can lead to different adaptation needs~\cite{11670150}. A rank sufficient for one client's task may be insufficient for another. Allowing client-specific ranks can fit both resource constraints and differences in local data~\cite{cho2024heterogeneous,bai2024federated}. We treat these heterogeneous ranks as given, and study the server operations needed to support them. Heterogeneous ranks introduce two distinct server tasks. Before aggregation, the server must handle uploaded adapters with different dimensions and scaling conventions. After aggregation, it must reconstruct an LoRA factor pair that fits each client's LoRA rank and send it to the client for the next local training round. We refer to these tasks as \textit{pre-aggregation alignment} and \textit{rank-specific LoRA adapter extraction}, respectively. Although client LoRA adapters differ in size, they represent updates to the same full-size weight matrix. If a client's LoRA rank is below the rank of the server's aggregated update, exact reconstruction is impossible. Therefore, some approximation error is unavoidable. The matrix information retained within that client rank also depends on how the smaller adapters are constructed. These operations recur in every communication round, so their computation and memory requirements affect the efficiency of the training procedure~\cite{bonawitz2019flsystem}. For a fixed aggregate, a direct approach to adapter extraction is to form the dense update matrix and apply singular value decomposition (SVD), as illustrate in Fig.\ref{fig:heterogeneous_redistribution}. Retaining its leading singular components minimizes Frobenius reconstruction error under a given rank constraint~\cite{eckart1936approximation}. However, applying SVD to a matrix with the same dimensions as the pretrained weight can require substantial computation time. Forming this matrix also adds computation and storage costs~\cite{liao2024apiq,meng2024pissa}. Repeating these operations across large adapted layers and training rounds can increase server processing costs. Direct prefix truncation of the original aggregated adapters avoids dense reconstruction and decomposition. However, the selected coordinates do not generally guarantee the minimum reconstruction error at the requested rank. These extraction choices are distinct from how the aggregate is constructed. We distinguish two aggregation forms. \textit{Product-Sum (PS)} aggregation averages the $B$ and $A$ factors separately and represents the aggregate by their product. \textit{Sum-Product (SP)} aggregation instead combines the paired client products as complete updates. Neither form prescribes a particular extraction method. FlexLoRA follows the SP formulation and applies SVD to the dense aggregate to construct rank-specific adapters~\cite{bai2024federated}. FLoRA also follows the SP formulation, but represents the weighted sum of client updates through stacked adapters~\cite{wang2024flora}. This avoids SVD, although the global adapter width equals the sum of participating client ranks. Broadcasting the stacked pair therefore sends more adapter entries than returning client-sized pairs. Methods based on PS aggregation handle heterogeneous adapter dimensions before averaging. HetLoRA pads smaller adapters before sparsity-weighted aggregation and returns truncated adapters~\cite{cho2024heterogeneous}. Its global adapter width is bounded by the maximum participating client rank. When a common set of client weights is used across positions, padded zeros can attenuate entries supplied only by higher-rank clients. Our prior RBLA method instead uses support-aware averaging to exclude missing positions~\cite{chen2024rbla}. It retains a compact aggregated pair but still constructs client adapters through direct prefix truncation. Therefore, its treatment of missing positions does not remove the potential excess reconstruction error introduced during extraction.

\begin{figure}[t]
  \centering
    \includegraphics[width=0.9\linewidth]{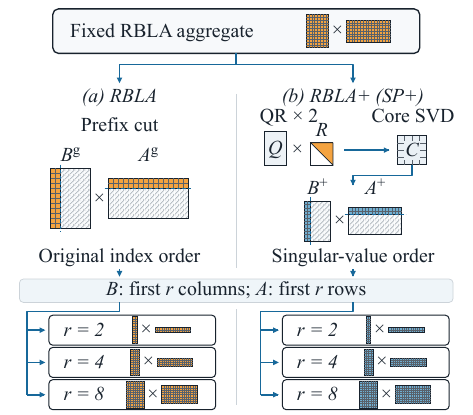}
  \caption{Two ways to extract client adapters from the same aggregate. (a) RBLA truncates the original factors. (b) RBLA+ constructs equivalent factors ordered by singular value using two QR factorizations and a small-core SVD. Both send the first $r$ columns of the left factor and the first $r$ rows of the right factor. Spectral extraction minimizes Frobenius reconstruction error at each client rank. Hatching marks omitted entries.}\vspace{-4 mm}
  \label{fig:0103}
\end{figure}

The limitation of prefix extraction is more general than a particular aggregation rule. Different LoRA factors can represent the same complete update, a non-uniqueness known as \textit{gauge freedom}. Their prefixes can nevertheless represent different lower-rank updates. Therefore, a rank-slot index alone provides no representation-independent guarantee of reconstruction importance. The resulting error can exceed the minimum imposed by the client's LoRA rank. We analyze this representation dependence and examine how it appears in the compact adapters produced by heterogeneous-rank aggregation. This separates a choice of representation from the unavoidable approximation required by a smaller client budget. We propose RBLA+ to combine compact aggregation with accurate rank-specific extraction. Pre-aggregation alignment handles adapter dimensions and converts uploaded adapters to the server's LoRA scaling convention. The aggregation stage retains RBLA's support-aware rule. The extraction stage, Spectral Projection Plus (SP+), uses established QR factorizations and a small-core SVD to construct ordered adapters without forming the dense update. Each client receives the prefix matching its rank, converted to its own scaling convention. Fig.~\ref{fig:0103} contrasts this construction with direct truncation. For the fixed aggregated update, SP+ minimizes Frobenius reconstruction error at every supported rank and preserves the complete update at the maximum rank~\cite{eckart1936approximation}. For a fixed maximum adapter width, compact decomposition costs scale linearly with the row and column dimensions of the adapted weight matrix.

RBLA+ preserves the assigned client adapter sizes, so its per-round adapter payload matches methods that transmit the same client-sized pairs. It avoids FLoRA's summed-rank downlink while retaining the per-round payload of matched-rank compact baselines. The guarantee applies to reconstruction of the chosen aggregate. Its consequences for task accuracy are evaluated empirically. We evaluate these practical effects using vision and language tasks, reconstruction experiments, and CPU benchmarks of adapter extraction. This journal extension studies the reconstruction loss when aggregated global LoRA update is reconstructed to client-specific LoRA adapters. It extends RBLA~\cite{chen2024rbla} with an analysis of representation-dependent prefix error, scale-consistent spectral extraction for all supported client ranks, and a broader evaluation of reconstruction behavior, task accuracy, and extraction costs.

\begin{itemize}
\item We analyze the representation dependence of adapter-prefix extraction. We show how equivalent LoRA adapter representations can produce different prefix reconstructions and identify conditions under which the reconstruction remains unchanged. The analysis distinguishes excess reconstruction error introduced by extraction from the unavoidable approximation error imposed by the client rank. We examine these effects using both equivalent representations and adapters produced by aggregation.

\item RBLA+ introduces two changes to the original RBLA: compact spectral extraction and bidirectional scale conversion. The first replaces direct prefix extraction with established QR/core-SVD decomposition. It attains the minimum Frobenius reconstruction error for the fixed aggregate at each supported rank without forming the dense update. The second accounts for client and server scaling conventions during upload and broadcast. Both modifications operate on compact adapters. The extension retains RBLA's support-aware aggregation rule, prescribed client adapter sizes, and per-round adapter payload.

\item We broaden the evaluation through heterogeneous vision and language tasks, controlled reconstruction experiments, ablations, and server cost measurements. Accuracy comparisons use different client rank configurations. Reconstruction and ablation studies examine the extraction mechanism, scale conversion, and their relationship with training behavior. CPU benchmarks with synthetic adapters of model-specific dimensions measure extraction time and additional process RSS. Analytical payload comparisons further characterize the communication costs of serving clients at their prescribed ranks.
\end{itemize}

\section{Related Work}
\label{sec:related-work}

\begin{table*}[!t]
\centering
\caption{Qualitative comparison of representative LoRA methods and systems with rank-specific adapter extraction in RBLA+.}
\label{tab:lora-landscape}

\renewcommand{\arraystretch}{1.12}
\setlength{\tabcolsep}{2.3pt}
\scriptsize

\begin{tabularx}{\textwidth}{
    @{}
    >{\RaggedRight\arraybackslash}p{0.14\textwidth}
    >{\RaggedRight\arraybackslash}p{0.22\textwidth}
    *{5}{c}
    >{\RaggedRight\arraybackslash}X
    @{}
}
\hline
\noalign{\vskip 3pt}

\makecell[l]{\textbf{Category}}
& \makecell[l]{\textbf{Representative methods}}
& \makecell[c]{\textbf{Het. rank}}
& \makecell[c]{\textbf{Collab.}}
& \makecell[c]{\textbf{Matrix}}
& \makecell[c]{\textbf{Multi-rank}}
& \makecell[c]{\textbf{Compact}}
& \makecell[l]{\textbf{Main difference from SP+}}
\\

\noalign{\vskip 3pt}
\hline
\noalign{\vskip 2pt}

LoRA formulations and rank adaptation
& LoRA, AdaLoRA, DyLoRA, QLoRA, LoRA+, DoRA, PiSSA, LoRA-GA
~\cite{hu2021lora,zhang2023adalora,valipour2023dylora,dettmers2023qlora,hayou2024loraplus,liu2024dora,meng2024pissa,wang2024loraga}
& \textasciitilde
& $\times$
& \textasciitilde
& \textasciitilde
& $\checkmark$
& Improves local rank allocation, memory, scaling, decomposition, or
initialization, but does not aggregate independent client adapters or define a
shared representation for client-specific extraction. \\
\hline

LoRA theory and matrix foundations
& Intrinsic dimension, LoRA expressivity, LoRA+, RAC-LoRA, LoRA-One, truncated
SVD~\cite{li2018intrinsic,aghajanyan2021intrinsic,zeng2024expressive,hayou2024loraplus,malinovsky2025rac,zhang2025loraone,eckart1936approximation}
& \textasciitilde
& \textasciitilde
& $\checkmark$
& $\times$
& \textasciitilde
& Provides expressivity, convergence, and approximation results, including a
federated RAC-LoRA extension. Rank-specific adapter extraction is not its
primary focus. \\
\hline

Federated LoRA with compatible adapters
& FedIT, pFedLoRA, FFA-LoRA, FedSA-LoRA, FLASC, LoRA-FAIR, RoLoRA
~\cite{zhang2023fedit,yi2024pfedlora,sun2024improving,guo2025selective,kuo2024flasc,bian2024lorafair,chen2025rolora}
& \textasciitilde
& $\checkmark$
& \textasciitilde
& $\times$
& $\checkmark$
& Studies collaborative tuning, communication, personalization, or aggregation
bias, but usually assumes compatible adapters and one broadcast rank. \\
\hline

Heterogeneous adapter-coordinate aggregation
& HetLoRA, replication, RBLA
~\cite{cho2024heterogeneous,byun2024towards,chen2024rbla}
& $\checkmark$
& $\checkmark$
& $\times$
& $\checkmark$
& \textasciitilde
& Supports heterogeneous ranks through adapter-coordinate aggregation, but
prefix-based extraction remains representation-dependent. \\
\hline

Update-/subspace-aware federated LoRA
& FLoRA, FlexLoRA, FLoRIST
~\cite{wang2024flora,bai2024federated,ramesh2025florist}
& $\checkmark$
& $\checkmark$
& $\checkmark$
& \textasciitilde
& \textasciitilde
& Constructs the global update from client matrices or subspaces, potentially
requiring a dense update, growing intermediate rank, or unified-rank
objective. \\
\hline

Distributed and split LoRA training
& SplitLoRA, mLoRA~\cite{lin2024splitlora,ye2025mlora}
& \textasciitilde
& \textasciitilde
& $\times$
& $\times$
& $\times$
& Optimizes partitioning, pipelines, operators, and accelerator utilization,
not heterogeneous aggregation or rank-specific adapter extraction. \\
\hline

Multi-tenant LoRA serving
& Punica, S-LoRA, dLoRA, Toppings
~\cite{chen2024punica,sheng2024slora,wu2024dlora,li2025toppings}
& \textasciitilde
& $\times$
& $\times$
& \textasciitilde
& $\times$
& Batches, places, pages, or migrates existing adapters. It handles runtime
rank diversity but does not construct a global adapter from training
updates. \\
\hline

Rank-specific extraction in RBLA+
& Compact QR/core-SVD decomposition after support-aware aggregation
& $\checkmark$
& $\checkmark$
& $\checkmark$
& $\checkmark$
& $\checkmark$
& Preserves a fixed aggregate and constructs nested approximations with
minimum Frobenius error at each supported adapter rank using compact
QR/core-SVD decomposition. \\
\hline
\end{tabularx}

\par\smallskip
\begin{minipage}{\textwidth}
\scriptsize
\RaggedRight
\textit{Note:}
Het. rank = explicit support for multiple LoRA ranks; Collab. = aggregation or coordinated training across data owners; Matrix = reasoning or transformation at the functional update-matrix/subspace level rather than relying only on raw adapter coordinates; Multi-rank = production of rank-specific models from a shared representation; Compact = no dense $d_{\mathrm{out}}\times d_{\mathrm{in}}$ server-side transformation.
$\checkmark$ indicates explicit support, $\times$ indicates no explicit
support, and \textasciitilde{} indicates partial, indirect, or
method-dependent support.
\end{minipage}

\end{table*}

\subsection{Low-Rank Adaptation and Rank-Adaptive Variants}

Parameter-efficient fine-tuning (PEFT) adapts a pretrained model by optimizing only a small set of parameters. LoRA represents the update of a pretrained weight matrix as the product of two thin trainable factors~\cite{hu2021lora}. Its motivation is consistent with earlier observations that highly over-parameterized models can often be optimized in much smaller intrinsic subspaces~\cite{li2018intrinsic,aghajanyan2021intrinsic}. A trained LoRA update can be merged into the pretrained weight, allowing inference without a separate adapter branch. Subsequent work improves different parts of the LoRA pipeline. AdaLoRA allocates the parameter budget across layers by pruning less important singular components~\cite{zhang2023adalora}, whereas DyLoRA trains nested prefixes so that one adapter can operate over a range of ranks without retraining~\cite{valipour2023dylora}. QLoRA combines low-rank updates with a frozen quantized backbone to reduce fine-tuning memory~\cite{dettmers2023qlora}. LoRA+ derives different learning rates for the two factors from large-width scaling arguments~\cite{hayou2024loraplus}, and DoRA separates weight magnitude from direction before applying a low-rank directional update~\cite{liu2024dora}. Spectral initialization methods such as PiSSA and LoRA-GA align the initial adapter with principal weight or gradient directions to improve optimization~\cite{meng2024pissa,wang2024loraga}. These methods show that rank allocation, adapter parameterization, and initialization can affect optimization and task performance. However, they primarily study a single optimization process and do not determine how independently trained adapters should be aggregated and redistributed across clients.

\subsection{Theoretical and Matrix Foundations of LoRA}

The empirical success of LoRA is commonly associated with the low intrinsic dimension of downstream adaptation. Intrinsic-dimension studies show that useful solutions can exist in subspaces far smaller than the full parameter space~\cite{li2018intrinsic,aghajanyan2021intrinsic}. Expressivity analysis further characterizes how the required LoRA rank depends on model width, depth, and the target function, and provides approximation bounds below the sufficient-rank threshold~\cite{zeng2024expressive}. Optimization-oriented analyses expose a separate issue: a low-rank factorization changes the geometry and dynamics of the optimization problem. LoRA+ analyzes factor-wise learning-rate scaling~\cite{hayou2024loraplus}. RAC-LoRA develops a randomized low-rank framework with convergence guarantees toward full-parameter solutions~\cite{malinovsky2025rac}. LoRA-One connects spectral initialization to the singular subspace of an initial full gradient~\cite{zhang2025loraone}. These results support low-dimensional adaptation but do not uniquely determine the LoRA adapter representing a given update. For an invertible matrix $G$, the pairs $(B,A)$ and $(BG,G^{-1}A)$ produce the same complete update. Consequently, a LoRA factor coordinate alone does not provide a representation-independent measure of reconstruction importance. At the matrix level, the Eckart-Young-Mirsky theorem gives a representation-independent answer to rank-constrained approximation: truncated SVD minimizes the Frobenius-norm reconstruction error~\cite{eckart1936approximation,golub2013matrix}. SP+ applies this classical result through established QR/core-SVD decomposition, obtaining rank-specific approximations without materializing the dense update~\cite{rblaMethodBallaniKressner2016}.

\subsection{Federated LoRA Fine-Tuning}

Federated learning coordinates optimization over decentralized data without collecting raw client records~\cite{mcmahan2017fedavg,kairouz2021advances}. FedIT demonstrated federated instruction tuning of language models on diverse private instruction data~\cite{zhang2023fedit}, while pFedLoRA uses a shared small adapter to exchange knowledge among heterogeneous personalized local models~\cite{yi2024pfedlora}. FLASC reduces communication by sparsifying transmitted LoRA updates while retaining dense local adapter training~\cite{kuo2024flasc}.

A central difficulty is that separately averaging the two LoRA factors does not generally yield the average of the complete client updates. FFA-LoRA freezes a shared factor $A$ and communicates only the trainable factor $B$, avoiding the mismatch caused by separately averaging both factors and reducing communication~\cite{sun2024improving}. FedSA-LoRA keeps both factors trainable but aggregates only a shared factor~\cite{guo2025selective}. LoRA-FAIR introduces aggregation and initialization corrections~\cite{bian2024lorafair}, whereas RoLoRA uses alternating optimization to improve robustness under data heterogeneity and small adapter budgets~\cite{chen2025rolora}. These methods address communication, personalization, initialization, or adapter-aggregation bias, but generally assume that participating clients exchange compatible adapter structures.

\subsection{Heterogeneous-Rank Federated LoRA}

Clients with different budgets can use different adapter ranks~\cite{liu2025adaptive,lin2026hsplitlora}. HetLoRA pads smaller adapters to a common shape for aggregation and truncates the resulting global adapters for redistribution~\cite{cho2024heterogeneous}. Replication-based padding fills missing positions in lower-rank adapters using corresponding components aggregated from high-rank clients~\cite{byun2024towards}. RBLA instead normalizes the aggregation weights over the clients contributing to each adapter position~\cite{chen2024rbla}. This avoids treating missing positions as zero-valued observations. These approaches address dimensional compatibility through compact adapter operations. Their lower-rank outputs nevertheless depend on the adapter coordinates selected for each client. Update-space methods preserve each client's paired factors before combining them. FLoRA stacks weighted client adapters so that their product represents a sum of complete local updates~\cite{wang2024flora}. FlexLoRA reconstructs a full-size aggregate and applies SVD to construct adapters at different client ranks~\cite{bai2024federated}. FLoRIST integrates stacked client adapters in a compact latent space and uses singular-value energy thresholds to choose a shared broadcast rank for each layer~\cite{ramesh2025florist}.

RBLA+ extracts client-specific adapters from a fixed compact aggregate. The product of separately averaged factors generally differs from the sum of client effective updates used by update-space methods. We preserve this distinction when interpreting reconstruction and accuracy results. Building on established compact QR/core-SVD decomposition~\cite{rblaMethodBallaniKressner2016}, SP+ constructs nested spectral prefixes for all requested ranks using one decomposition. This removes excess Frobenius reconstruction error for the chosen aggregate without forming the dense update or increasing the assigned client adapter sizes.

\subsubsection{Relationship to our prior RBLA work}
Our conference paper introduced support-aware adapter aggregation and prefix-based redistribution for heterogeneous ranks~\cite{chen2024rbla}. The present work retains that aggregation rule. Its new focus is the information lost when an aggregated representation is reduced to a client rank. We analyze representation-dependent prefix error, replace direct truncation with compact spectral extraction, and incorporate bidirectional scale conversion. The evaluation examines reconstruction mechanisms, training behavior, and the computation and memory required to prepare client adapters. QR/core-SVD decomposition is an established numerical technique used in this design, rather than a new matrix decomposition~\cite{rblaMethodBallaniKressner2016}.

\subsection{Distributed Training and Split Fine-Tuning}

LoRA also appears in distributed systems that focus on model execution rather than adapter aggregation and extraction. SplitLoRA combines LoRA with split federated learning to offload large-model computation and exchange activations across a model cut~\cite{lin2024splitlora}. mLoRA introduces LoRA-aware pipeline parallelism and specialized operators for concurrently training multiple adapters across GPUs and machines~\cite{ye2025mlora}. These systems focus on placement, pipelining, and resource utilization. They do not construct a global adapter from independently optimized heterogeneous-rank adapters. Multi-tenant serving systems exploit the fact that many adapters share one base model. Punica batches operations from different adapters with specialized GPU kernels and cluster scheduling~\cite{chen2024punica}. S-LoRA combines unified paging, heterogeneous batching, and tensor parallelism to serve many adapters, including adapters with different ranks~\cite{sheng2024slora}. dLoRA dynamically merges, migrates, and co-schedules requests and adapters across replicas~\cite{wu2024dlora}, while Toppings uses CPU assistance and rank-aware scheduling to reduce adapter-loading and padding overheads~\cite{li2025toppings}. These systems address rank heterogeneity during inference over existing adapters, whereas our focus is rank-specific reconstruction of a fixed aggregated update at the server. Table~\ref{tab:lora-landscape} summarizes the resulting research landscape. The comparison distinguishes heterogeneous-rank support from collaborative training, update-matrix or subspace processing, rank-specific extraction, and compact server-side transformation.

\section{Background and Preliminaries}
\label{sec:preliminaries}

\begin{table}[t]
\centering
\caption{Main notation for the mathematical analysis.}
\label{tab:notation}
\begingroup
\footnotesize
\setlength{\tabcolsep}{3pt}
\renewcommand{\arraystretch}{1.08}
\begin{tabularx}{\columnwidth}{@{}l>{\raggedright\arraybackslash}X@{}}
\toprule
Notation & Meaning \\
\midrule
$d_{\mathrm{out}},d_{\mathrm{in}}$
& Output and input dimensions. \\
$r_i,R$
& Client rank and maximum client rank. \\
$W_0$
& Frozen pretrained weight matrix. \\
$B_i,A_i$
& Client $i$'s LoRA factors. \\
$B^g,A^g$
& Aggregated server LoRA factors. \\
$B^+,A^+$
& Spectrally ordered SP+ factors. \\
$B_i^{\mathrm{next}},A_i^{\mathrm{next}}$
& LoRA factors received for the next round. \\
$b_{i,s},a_{i,s}^{\mathsf T}$
& Column $s$ of $B_i$ and row $s$ of $A_i$. \\
$b_s^g,(a_s^g)^{\mathsf T}$
& Corresponding aggregated column and row. \\
\midrule
$M^g$
& Unscaled aggregate $B^gA^g$. \\
$M_r^{\mathrm{pre}},M_r^+$
& Prefix products of RBLA and SP+ at LoRA rank $r$. \\
$\widehat M_r$
& Reconstruction of rank at most $r$. \\
$s_i,s_s$
& Client and server LoRA scales. \\
$\Delta W_i,\Delta W^g$
& Effective updates $s_iB_iA_i$ and $s_sM^g$. \\
$G,\Pi_r$
& Invertible basis change and prefix selector, respectively. \\
\midrule
$E_r$
& Frobenius reconstruction error. \\
$E_r^\star$
& Minimum error at LoRA rank $r$. \\
$\varepsilon_r$
& Normalized excess error
$(E_r-E_r^\star)/\|M^g\|_F$. \\
\bottomrule
\end{tabularx}
\par\smallskip
\endgroup
\end{table}

We consider a server coordinating federated fine-tuning among clients with different LoRA adapter ranks. Clients keep their training data local and exchange LoRA adapters after an initial distribution of the pretrained model. We take each client's LoRA rank as a prescribed input and focus on rank-specific adapter extraction and broadcast. Our objective is to preserve the aggregated update as accurately as possible within these rank budgets while keeping server-side processing costs low. We first define the client representation and server operations for one adapted linear layer in a single communication round. The reconstruction analysis conditions on a fixed aggregate, independently of the local optimization procedure that produced the uploaded adapters. Table~\ref{tab:notation} summarizes the main notation.

\subsection{Heterogeneous-Rank LoRA}

For a frozen pretrained weight matrix $W_0\in\mathbb{R}^{d_{\mathrm{out}}\times d_{\mathrm{in}}}$, client $i$ maintains LoRA factors
\begin{equation}
    B_i\in\mathbb{R}^{d_{\mathrm{out}}\times r_i},
    \qquad
    A_i\in\mathbb{R}^{r_i\times d_{\mathrm{in}}},
    \label{eq:prelim-local-factors}
\end{equation}
whose scaled product defines the effective update:
\begin{equation}
    \Delta W_i=s_iB_iA_i.
    \label{eq:prelim-local-update}
\end{equation}
Here, $B_iA_i$ is the unscaled factor product, and $s_i>0$ is the layer's scaling coefficient. Therefore, the effective weight is $W_0+\Delta W_i$. Adapter dimensions determine how many parameters a client trains and exchanges. The scaling coefficient determines how their product changes the pretrained weight. These two properties must be handled separately when clients use different configurations. Let $N$ be the number of participating clients, with ranks $1\le r_i\le R$, where $R=\max_i r_i$ and $R\le\min\{d_{\mathrm{out}},d_{\mathrm{in}}\}$. The LoRA ranks are given to the server and remain unchanged by adapter extraction. Client aggregation weights satisfy $p_i>0$ and $\sum_{i=1}^{N}p_i=1$. Writing $b_{i,s}\in\mathbb{R}^{d_{\mathrm{out}}}$ for column $s$ of $B_i$ and $a_{i,s}^{\mathsf T}$ for row $s$ of $A_i$, with $a_{i,s}\in\mathbb{R}^{d_{\mathrm{in}}}$, gives
\begin{equation}
    B_iA_i=\sum_{s=1}^{r_i}b_{i,s}a_{i,s}^{\mathsf T}.
    \label{eq:prelim-rank-one}
\end{equation}
Here, the index $s$ identifies an adapter position and need not reflect the singular-value order of $B_iA_i$. For this linear layer, client $i$ trains and exchanges $(d_{\mathrm{out}}+d_{\mathrm{in}})r_i$ adapter parameters per direction. A given rank thus specifies an adapter parameter budget and, at fixed precision, a communication payload. It does not specify total training memory, which also includes the backbone and other training state.

\subsection{Aggregation and Adapter Extraction}

Aggregation constructs a global representation from uploaded client adapters. Rank-specific adapter extraction constructs a pair at each recipient's prescribed rank. Broadcast then transmits the resulting pair. These are separate operations even when an implementation combines them. Let $p_i>0$ denote the aggregation weight assigned to
client $i$, at the unscaled product level, Sum-Product (SP) aggregation is written as:
\begin{equation}
    \sum_{i=1}^{N}p_iB_iA_i.
    \label{eq:prelim-sp}
\end{equation}
Product-Sum (PS) methods instead aggregate the two factors separately and represent the result by their product. These names describe the aggregation expressions, not mandatory implementations or extraction rules. A matrix-level expression may be maintained implicitly. Likewise, adapter-level aggregation does not require direct prefix extraction. FlexLoRA uses matrix-level aggregation followed by SVD-based adapter construction~\cite{bai2024federated}. RBLA uses adapter-level aggregation followed by direct prefix extraction~\cite{chen2024rbla}.

\subsection{Support-Aware Aggregation in RBLA}

We retain RBLA's support-aware aggregation~\cite{chen2024rbla} to keep the aggregate representation bounded by the maximum participating adapter width. For each adapter index $s$, it includes only clients whose adapters contain that index. Its participating set and normalized weights are:
\begin{align}
    \mathcal{I}_s
    &=\{i:r_i\ge s\},
    \label{eq:prelim-rbla-participation}\\
    \alpha_{i,s}
    &=\frac{p_i}{\sum_{j\in\mathcal{I}_s}p_j},
    \qquad i\in\mathcal{I}_s.
    \label{eq:prelim-rbla-weight}
\end{align}
The denominator is positive because $p_i>0$ and $\mathcal{I}_s$ is nonempty for $s\le R$. Unlike zero padding with global client weights, RBLA renormalizes the weights at each index over the clients whose adapters include that index. Under equal client weights, $\alpha_{i,s}=1/|\mathcal{I}_s|$. At higher adapter indices, only clients with sufficiently large ranks contribute. At an index absent from some clients, the remaining clients receive normalized weights greater than their original global weights $p_i$. The corresponding columns and rows are averaged separately:
\begin{equation}
    b_s^g=\sum_{i\in\mathcal{I}_s}\alpha_{i,s}b_{i,s},
    \qquad
    a_s^g=\sum_{i\in\mathcal{I}_s}\alpha_{i,s}a_{i,s}.
    \label{eq:prelim-rbla-aggregation}
\end{equation}
Stacking these vectors gives $B^g=[b_1^g,\ldots,b_R^g]$ and $A^g=[a_1^g,\ldots,a_R^g]^{\mathsf T}$, with aggregate product
\begin{equation}
    M^g=B^gA^g=\sum_{s=1}^{R}b_s^g(a_s^g)^{\mathsf T}.
    \label{eq:prelim-rbla-update}
\end{equation}
Separate factor averaging introduces cross-client products:
\begin{equation}
    b_s^g(a_s^g)^{\mathsf T}
    =\sum_{i,j\in\mathcal{I}_s}
    \alpha_{i,s}\alpha_{j,s}
    b_{i,s}a_{j,s}^{\mathsf T}.
    \label{eq:prelim-rbla-cross-products}
\end{equation}
These normalized weights also determine the cross-client coefficients $\alpha_{i,s}\alpha_{j,s}$ in Eq.~\eqref{eq:prelim-rbla-cross-products}, which can vary across adapter indices. Therefore, $B^gA^g$ generally differs from \eqref{eq:prelim-sp}. These equations describe aggregation of the supplied adapters. RBLA+ applies the same rule after converting the uploads to the server scaling convention in Section~\ref{subsec:rank-scale-alignment}. Its extraction stage then operates on the resulting product. For a target rank $r\le R$, original RBLA retains the first $r$ columns of $B^g$ and rows of $A^g$, producing
\begin{equation}
    M_r^{\mathrm{pre}}=B^g_{:,1:r}A^g_{1:r,:}.
    \label{eq:prelim-rbla-prefix-update}
\end{equation}
Both operations use compact adapters. The inner dimension $R$ is a representation budget, and the actual rank of $M^g$ may be smaller. The next section asks whether a direct prefix uses a smaller client budget effectively.

\section{Problem Statement}
\label{sec:problem-statement}
In federated fine-tuning with heterogeneous LoRA ranks, the server must extract from an aggregated update an adapter that fits each client's rank. Approximation error is unavoidable when a client's rank is below the rank of the aggregated update, and the extraction rule may add more error. For example, direct prefix extraction keeps the first $r$ columns of the left factor and the first $r$ rows of the right factor, although these positions need not reflect their importance for reconstruction. As a result, different factor pairs of the same update can give different prefix reconstructions at the same rank. Reconstruction quality thus depends on the factor representation, even when the aggregate and the client rank are fixed. This raises a general question: how much error comes from the rank constraint, and how much comes from the representation and extraction rule? We first identify when equivalent representations preserve or change prefix reconstructions, and then examine the resulting excess error in aggregation outputs.

\begin{figure}[t]
\centering
\includegraphics[width=1\linewidth]{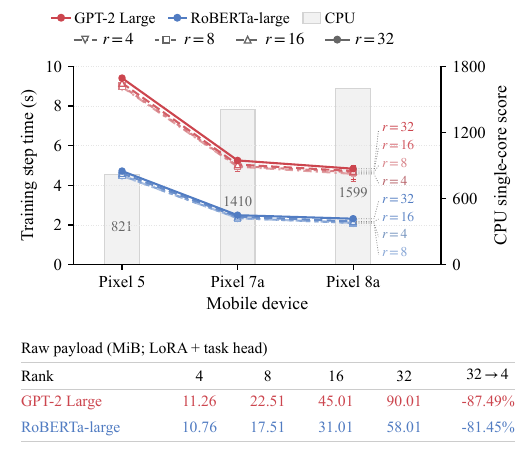}
\caption{Training latency and parameter payload at different LoRA ranks. Curves show GPT-2 Large (red) and RoBERTa-large (blue) training-step latency on three Pixel smartphones at $r\in\{4,8,16,32\}$.  Gray bars show public Geekbench~6 single-core reference scores. The table reports the raw FP32 payload of one copy of the LoRA factors and task head, in MiB, and its reduction from rank 32 to rank 4.}
\label{fig:mobile_problem_statement}
\end{figure}

This question matters in practice because clients have different communication budgets. On three Pixel smartphones (FP32, batch size 1, sequence length 64, SGD, one CPU thread), reducing the rank from 32 to 4 lowers the raw parameter payload, including the task head, by 87.49\% for GPT-2 Large and 81.45\% for RoBERTa-large, whereas the median training-step latency drops by only 4.74\%--9.22\% across the tested model--device pairs (Fig.~\ref{fig:mobile_problem_statement}). In this setting, rank mainly changes the communication payload rather than the step time. Clients with ample budgets can afford larger ranks, whereas clients with tight budgets may still need smaller adapters. Since adapters of all ranks are derived from the same aggregate, the question above must be answered at every prescribed rank at once, with each client's payload kept fixed. Truncated SVD gives the minimum Frobenius error at each rank~\cite{eckart1936approximation}, but forming and decomposing the dense update adds server computation and storage costs. We therefore seek a shared compact factor pair that preserves the full aggregate before truncation and whose prefixes give the minimum reconstruction error at every prescribed rank, without building or decomposing the dense update.

\subsection{Rank Constraints and Excess Reconstruction Error}
After aggregation, the server has a compact factor pair $B^g,A^g$ but must return smaller pairs to clients with lower LoRA ranks. We study extraction from the fixed product $M^g=B^gA^g$. The analysis applies to any pair of the stated dimensions. The implementation and training experiments use RBLA aggregation. Our objective is to preserve this fixed product within each client budget while keeping low server processing overhead. We define the adapter-index selector:

\begin{equation}
    \Pi_r=\operatorname{diag}(I_r,0_{R-r})
    \in\mathbb{R}^{R\times R}.
    \label{eq:prefix-projection}
\end{equation}
Direct prefix extraction gives $M_r^{\mathrm{pre}}=B^g\Pi_rA^g$, whose rank is at most $r$. To distinguish loss imposed by the LoRA rank from additional loss introduced by adapter extraction, we compare each reconstruction with the best approximation at the same rank. For $1\le r\le R$ and a reconstructed update $\widehat M_r$ of rank at most $r$, we measure reconstruction error using the unsquared Frobenius norm:
\begin{equation}
E_r=\lVert M^g-\widehat M_r\rVert_F.
\label{eq:reconstruction-error}
\end{equation}
Let $\sigma_1(M^g)\ge\cdots\ge\sigma_R(M^g)\ge0$, with zeros included when $\operatorname{rank}(M^g)<R$. By the Eckart-Young theorem~\cite{eckart1936approximation}, the minimum error under the same LoRA rank is
\begin{equation}
\begin{aligned}
E_r^\star
&=\min_{\operatorname{rank}(Z)\le r}\lVert M^g-Z\rVert_F \\
&=\left(\sum_{j=r+1}^{R}\sigma_j(M^g)^2\right)^{1/2}.
\end{aligned}
\label{eq:optimal-rank-r-error}
\end{equation}
This is the unavoidable approximation error for the fixed $M^g$ and LoRA rank. It is zero when $r\ge\operatorname{rank}(M^g)$.

For $M^g\ne0$, we define the normalized excess reconstruction error as
\begin{equation}
\varepsilon_r
=\frac{E_r-E_r^\star}{\lVert M^g\rVert_F}
\ge0.
\label{eq:normalized-excess-error}
\end{equation}
For $M^g=0$, we instead define $\varepsilon_r=E_r-E_r^\star$. Direct prefix extraction, as used by original RBLA, is evaluated by setting $\widehat M_r=M_r^{\mathrm{pre}}$ in Eq.~\eqref{eq:reconstruction-error}. Its excess error can depend on the adapter representation, whereas $E_r^\star$ depends only on $M^g$ and $r$. An appropriately ordered adapter representation can already have zero excess error. Prefix truncation does not guarantee this property for general aggregated adapters.

\subsection{Representation Dependence of Adapter Prefixes}

\begin{figure}[b]
\centering
\includegraphics[width=0.95\linewidth]{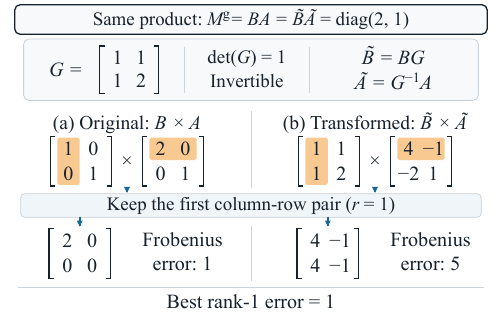}
\caption{Gauge transformation of LoRA adapters. The illustrated transformation $(B,A)\mapsto(BG,G^{-1}A)$ preserves $M^g=BA=\operatorname{diag}(2,1)$ but changes the rank-one prefix reconstruction. Its Frobenius error increases from $1$ to $5$, while the minimum achievable error remains $1$.}
\label{fig:0101}
\end{figure}

The adapter representation of a LoRA update is not unique. For any invertible $G\in\mathbb{R}^{R\times R}$, the transformed factors $B^gG$ and $G^{-1}A^g$ satisfy 
\begin{equation}
    (B^gG)(G^{-1}A^g)=B^gA^g=M^g.
    \label{eq:complete-invariance}
\end{equation}
We call this invertible change of basis in the internal rank coordinates a gauge transformation. Fig.~\ref{fig:0101} illustrates why this freedom matters for prefix truncation. Both LoRA factors represent the same aggregate product $M^g=\operatorname{diag}(2,1)$. Retaining only the first column of each left factor and the first row of its corresponding right factor, however, produces different rank-one reconstructions. In this example, the Frobenius reconstruction error increases from $1$ to $5$ after the transformation, while the minimum achievable rank-one error remains $1$. The figure thus shows how adapter-prefix selection can introduce avoidable error even when the complete update and the LoRA rank are unchanged.

\begin{figure}[t]
    \centering
    \begin{subfigure}[t]{0.24\textwidth}
        \centering
        \includegraphics[width=\linewidth]
        {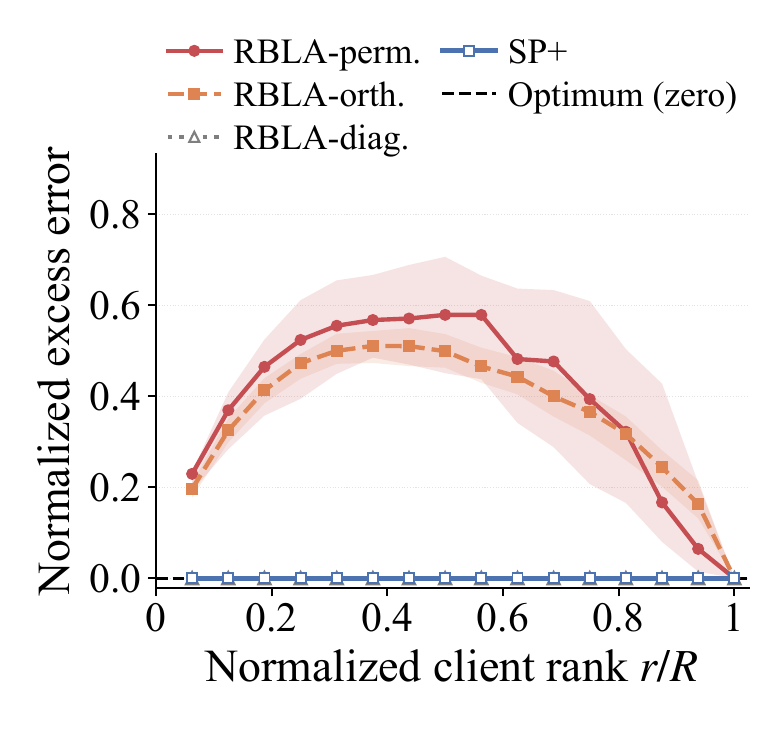}
        \caption{Gauge sensitivity.}
        \label{fig:gauge_excess_error}
    \end{subfigure}
    \hfill
    \begin{subfigure}[t]{0.24\textwidth}
        \centering
        \includegraphics[width=\linewidth]
        {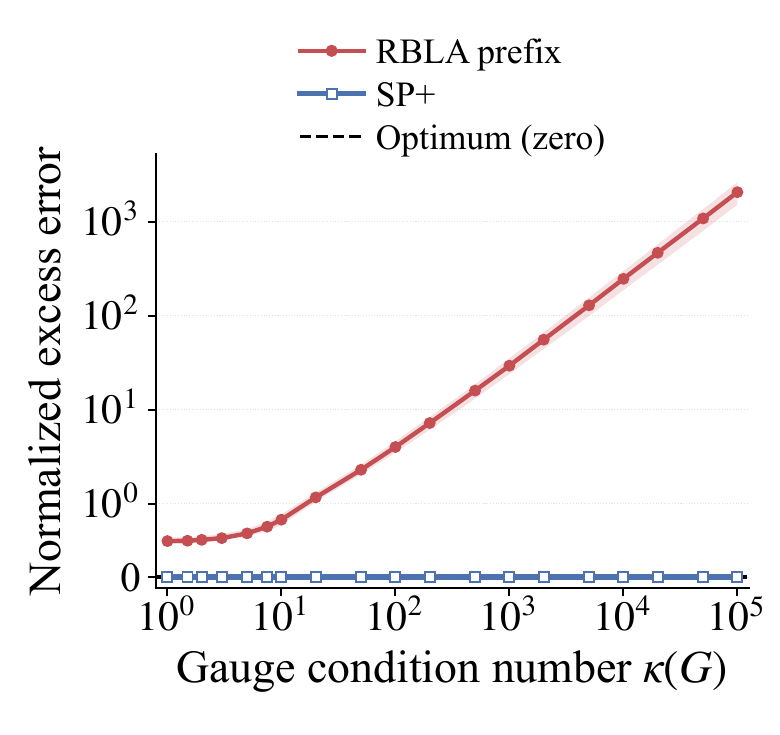}
        \caption{Gauge conditioning.}
        \label{fig:condition_number_excess_error}
    \end{subfigure}
    
    \begin{subfigure}[t]{0.24\textwidth}
        \centering
        \includegraphics[width=\linewidth]
        {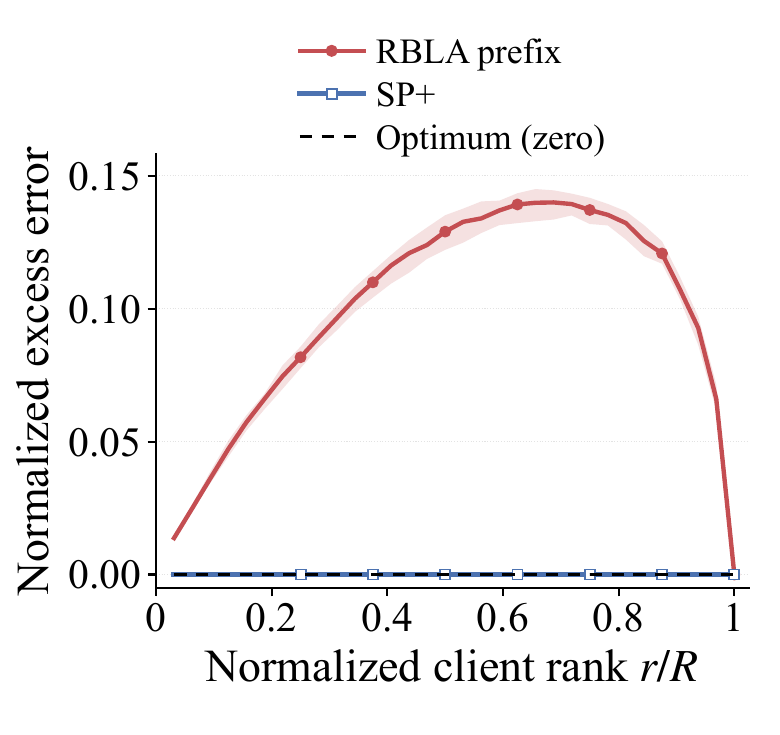}
        \caption{Excess reconstruction error.}
        \label{fig:rbla_excess_error}
    \end{subfigure}
    \hfill
    \begin{subfigure}[t]{0.24\textwidth}
        \centering
        \includegraphics[width=\linewidth]
        {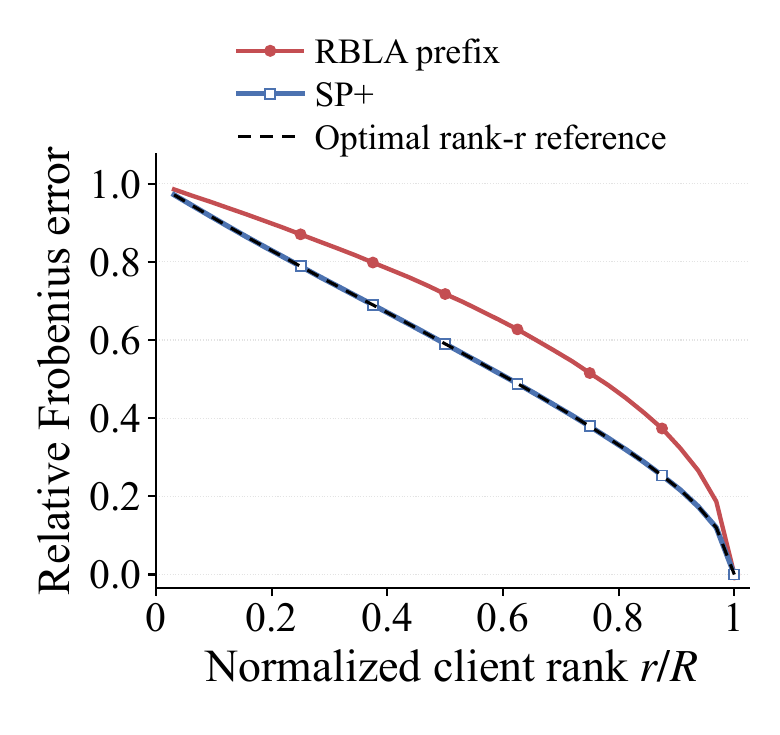}
        \caption{Relative reconstruction error.}
        \label{fig:rbla_relative_error}
    \end{subfigure}

    \caption{Reconstruction errors of direct prefixes and SP+. (a) Sensitivity to equivalent adapter representations. (b) Effect of conditioning within the tested gauge family. (c) Normalized excess error and (d) Relative error for the same synthetic RBLA aggregates, without additional gauge transformations. Curves and bands show medians and interquartile ranges. Black dashed lines indicate the optimal errors.}
\label{fig:rbla_reconstruction_analysis}\vspace{-4mm}
\end{figure}

More generally, applying the same index selector to the transformed factors produces
\begin{equation}
    B^gG\Pi_rG^{-1}A^g,
    \label{eq:transformed-prefix}
\end{equation}
which differs from $M_r^{\mathrm{pre}}$ if and only if $B^g(G\Pi_rG^{-1}-\Pi_r)A^g\ne0$. Therefore, an adapter index provides no general guarantee of reconstruction importance. If $B^g$ has full column rank and $A^g$ has full row rank, this condition is equivalent to $G\Pi_r\ne\Pi_rG$. Without these rank assumptions, noncommutation alone is insufficient. The effect depends on the transformation. If $G\Pi_r=\Pi_rG$, the prefix is unchanged. This includes diagonal rescaling and transformations acting separately within the retained and discarded coordinate blocks. Transformations mixing coordinates across the truncation boundary can change the prefix update, as illustrated in Fig.~\ref{fig:0101}. Diagonal rescaling preserves the original prefix and its reconstruction error. Whether that prefix is optimal depends on the original adapter representation. Fig.~\ref{fig:gauge_excess_error} examines this representation dependence using permutation, orthogonal, and diagonal transformations. Fig.~\ref{fig:condition_number_excess_error} further examines variation within the tested gauge family as $\kappa_2(G)=\sigma_{\max}(G)/\sigma_{\min}(G)$ changes. These experiments quantify the behavior illustrated by the example, while the condition number alone does not determine prefix reconstruction quality. Fig.~\ref{fig:condition_number_excess_error} and Fig.\ref{fig:gauge_excess_error} first examine whether equivalent representations of the same update yield different prefix errors. We use $128\times96$ synthetic updates with $R=16$. Fig.~\ref{fig:gauge_excess_error} evaluates 100 trials for each permutation, orthogonal, and diagonal gauge family. Permutation and orthogonal transformations produce positive excess error at intermediate ranks, whereas diagonal rescaling leaves each prefix unchanged. Fig.~\ref{fig:condition_number_excess_error} uses 100 paired trials at $r=8$ across condition-number levels $\kappa_2(G)=\sigma_{\max}(G)/\sigma_{\min}(G)$. The observed increase in error characterizes the tested family. The condition number alone does not determine prefix quality. 

\begin{figure}[t]
  \centering
    \includegraphics[width=\linewidth]{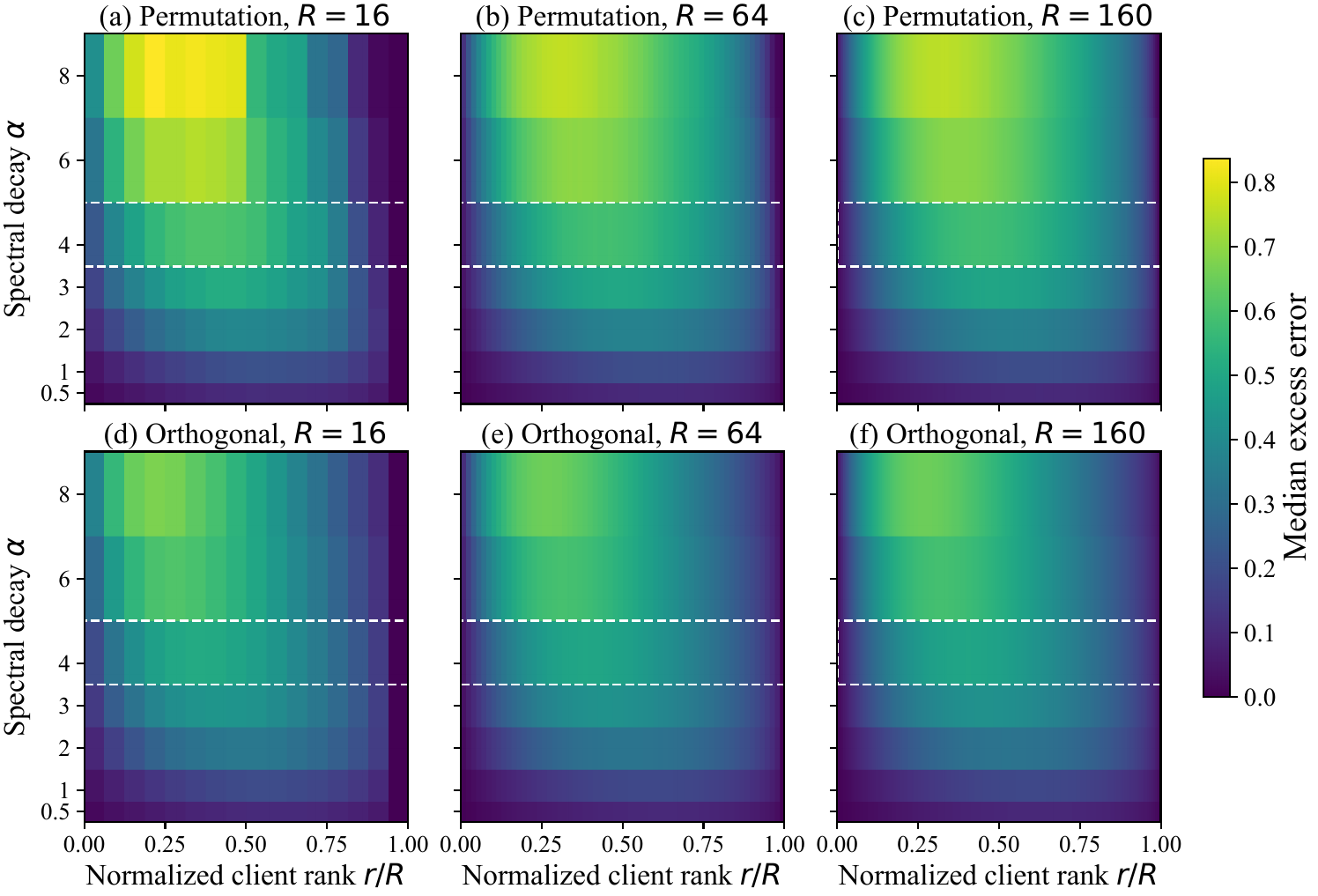}
  \caption{Normalized excess error $\varepsilon_r$ of index-prefix extraction across spectra and adapter widths. Rows correspond to permutation and orthogonal gauges. Columns correspond to $R\in\{16,64,160\}$. Colors show median errors, and white dashed outlines mark $\alpha=4$, the reference spectral decay.}\vspace{-4 mm}
  \label{fig:spectrum_robustness_heatmap}
\end{figure}

Fig.~\ref{fig:spectrum_robustness_heatmap} examines whether this representation sensitivity persists across different spectra and adapter widths $R\in\{16,64,160\}$. We use $\alpha\in\{0.5,1,2,3,4,6,8\}$ and singular values $\sigma_j(M^g)\propto\exp[-\alpha(j-1)/(R-1)]$, normalized so that $\sum_{j=1}^{R}\sigma_j(M^g)^2=1$. Larger $\alpha$ concentrates the update in fewer leading singular directions. Each synthetic update starts from balanced, singular-value-ordered factors, whose prefixes attain the optimal errors before transformation. For each $R$ and gauge family, the same 1,000 sampled transformations are reused across all spectral settings and target ranks $r=1,\ldots,R$. Each cell reports the median $\varepsilon_r$ for the transformed prefixes, with $E_r^\star$ computed from the spectral tail. The dashed $\alpha=4$ row matches the reference spectrum in Fig.~\ref{fig:gauge_excess_error}. The heatmaps show that the excess error persists across the tested adapter widths. At $\alpha=4$ and $r/R=0.5$, the median excess errors for permutation and orthogonal transformations are $0.592$ and $0.490$ at $R=16$, and remain $0.574$ and $0.477$ at $R=160$. The gap also depends on the spectrum: for orthogonal transformations at $R=160$ and $r/R=0.5$, it increases from $0.090$ at $\alpha=0.5$ to $0.542$ at $\alpha=8$ in this sweep. Both transformation families have $\kappa_2(G)=1$, so these gaps do not require ill-conditioned transformations. This experiment extends the evidence for representation sensitivity beyond a single spectrum and small adapter width. Figs.~\ref{fig:rbla_excess_error} and~\ref{fig:rbla_relative_error} then examine whether this reconstruction gap also occurs directly in RBLA aggregation outputs, without an additional gauge transformation. We use $N=10$ equally weighted synthetic clients, $d_{\mathrm{out}}=d_{\mathrm{in}}=512$, $R=32$, and client ranks $[8,8,12,12,16,16,20,24,28,32]$. Client adapters are generated from shared random reference adapters with relative perturbation scale $\eta=0.3$ over 30 independent seeds. Both extraction methods use the same aggregated LoRA adapter. Fig.~\ref{fig:rbla_excess_error} reports $\varepsilon_r$, isolating the error beyond the rank-constrained optimum, while Fig.~\ref{fig:rbla_relative_error} reports the total relative error $E_r/\lVert M^g\rVert_F$. The original prefixes have positive excess error at intermediate ranks, whereas SP+ matches the optimal reference within numerical precision. Thus, the reconstruction gap in this setting is present in the aggregation outputs themselves. Together, these controlled experiments show that prefix selection can introduce avoidable reconstruction loss, both under equivalent adapter representations and in untransformed synthetic RBLA aggregation outputs. They motivate an extraction rule that retains as much of the fixed aggregate as each client rank permits. The Frobenius criterion measures preservation of that aggregate without requiring client data.

\subsection{Compact Extraction for Multiple Client Ranks}

The server should obtain a shared LoRA factor pair $B^+\in\mathbb{R}^{d_{\mathrm{out}}\times R}$ and $A^+\in\mathbb{R}^{R\times d_{\mathrm{in}}}$. Writing $M_r^+=B^+_{:,1:r}A^+_{1:r,:}$ for its prefix product at LoRA rank $r$, the factors should satisfy
\begin{equation}
    \begin{aligned}
        B^+A^+&=M^g,\\
        \bigl\|M^g-M_r^+\bigr\|_F
        &=E_r^\star,\qquad 1\le r\le R.
    \end{aligned}
    \label{eq:redistribution-design-objective}
\end{equation}
These requirements preserve the complete aggregate and eliminate excess reconstruction error at every supported rank. A common representation should serve all clients through rank-specific slices, with compact intermediate storage and a client payload proportional to $(d_{\mathrm{out}}+d_{\mathrm{in}})r_i$. The optimality requirement concerns reconstruction error, rather than uniqueness of the adapters. At a repeated positive singular value across the truncation boundary, multiple optimal update matrices may exist. The minimum error in Eq.~\eqref{eq:redistribution-design-objective} remains well defined even when the optimal update matrix is not unique. RBLA+ addresses this extraction objective while retaining the given LoRA ranks and RBLA aggregation rule. SP+ preserves the aggregated product before rank truncation. Its extraction step approximates this fixed product without explicitly removing the cross-client terms introduced by RBLA aggregation. The next section integrates compact SP+ extraction into the complete upload, aggregation, and broadcast procedure.

\section{The RBLA+ Method}
\label{sec:proposed-method}

RBLA+ coordinates clients with prescribed adapter ranks through pre-aggregation alignment, support-aware aggregation, and rank-specific LoRA adapter extraction. Alignment handles compatible dimensions and scaling conventions. Aggregation retains the rule defined in Section~\ref{sec:preliminaries}. Extraction constructs client adapters from the resulting fixed aggregate, followed by scale conversion and broadcast. The local training procedure and configured client ranks are retained. For the linear layer considered above, the effective client update is
\begin{equation}
\Delta W_i=s_iB_iA_i.
\label{eq:rbla-plus-effective-update}
\end{equation}

The product $B_iA_i$ has rank at most $r_i$. A common convention is $s_i=\alpha_i/r_i$, where $\alpha_i$ is client $i$'s LoRA scaling numerator. But RBLA+ uses the actual module coefficient and does not assume equal $\alpha_i$ across clients.

\subsection{Pre-aggregation Alignment and Aggregation}
\label{subsec:rank-scale-alignment}
\label{subsec:tmc-support-aggregation}

\subsubsection*{Dimension alignment}
Different client ranks produce different adapter widths. The server organizes the uploaded adapters over a common width $R$. The support sets $\mathcal I_s$ in Eq.~\eqref{eq:prelim-rbla-participation} identify the clients contributing to each position. This operation can use the available columns and rows directly, without explicitly padding every upload or treating absent entries as observations. It provides dimensional compatibility but does not increase update rank or align the clients' internal representation bases.

\subsubsection*{Upload scale conversion}
Dimension alignment alone does not ensure consistent scaling. Client $i$ uses a LoRA coefficient $s_i>0$, while the server uses $s_s>0$. Before aggregation, RBLA+ converts each uploaded LoRA factor pair to the server's scaling convention. Scale compensation is also considered in FlexLoRA~\cite{bai2024federated}. Here, the conversion applies the same multiplier to both LoRA factors:

\begin{equation}
\begin{gathered}
c_i=\sqrt{\frac{s_i}{s_s}},\\
(B_i,A_i)\mapsto(c_iB_i,c_iA_i).
\end{gathered}
\label{eq:rank-scale-upload}
\end{equation}
This preserves the client's effective update under the server coefficient:
\begin{equation}
s_s(c_iB_i)(c_iA_i)=s_iB_iA_i.
\label{eq:rank-scale-upload-preservation}
\end{equation}
The operation converts scaling conventions rather than applying a gauge transformation to the unscaled product. The conversion depends on the actual scaling coefficients, not on adapter width alone. When $s_i=s_s$, it is the identity. If both sides use $s=\alpha/r$ with the same $\alpha$, then $c_i=\sqrt{r_s/r_i}$, where $r_s$ is the server module's configured rank. This configured rank need not equal the aggregate width $R$.

\subsubsection{Support-aware aggregation}
The server applies Eqs.~\eqref{eq:prelim-rbla-weight} and~\eqref{eq:prelim-rbla-aggregation} to the converted factors $(c_iB_i,c_iA_i)$. This retains RBLA's support sets and per-position weight normalization and produces an aggregated pair $B^g,A^g$ of width $R$. The aggregation rule is unchanged, while its inputs change when client and server scaling coefficients differ. Although conversion preserves each uploaded effective update, separate factor averaging still introduces cross-client products. Therefore, the server's effective update $s_sB^gA^g$ generally differs from $\sum_{i=1}^{N}p_i s_iB_iA_i$. The subsequent extraction stage preserves or approximates this chosen aggregate within each client's rank budget.

\subsection{SP+ Rank-Specific LoRA Adapter Extraction}
\label{subsec:spplus-broadcast}
\label{subsec:rank-specific-extraction}

Consider the aggregated pair $B^g\in\mathbb{R}^{d_{\mathrm{out}}\times R}$ and $A^g\in\mathbb{R}^{R\times d_{\mathrm{in}}}$. The following derivation uses $1\leq r_i\leq R\leq\min(d_{\mathrm{out}},d_{\mathrm{in}})$. SP+ handles wider adapters through a rectangular core and restores the configured output widths when necessary. We distinguish the factor product from the effective update:
\begin{equation}
M^g=B^gA^g,\qquad
\Delta W^g=s_sM^g.
\label{eq:spplus-server-update}
\end{equation}

\begin{figure}[b]
  \centering
  \includegraphics[width=\linewidth]{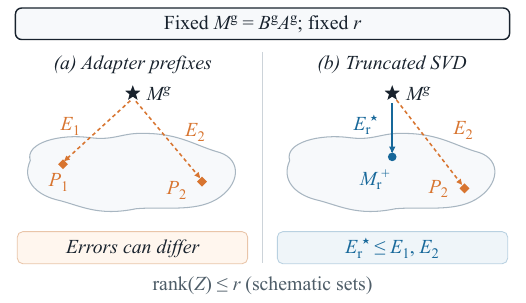}
  \caption{Schematic approximation of a fixed aggregate $M^g$ under LoRA rank $r$. (a) Equivalent adapter representations can yield different prefix reconstructions and errors. (b) Truncated SVD attains the minimum Frobenius error $E_r^\star$. The shaded regions represent matrices of rank at most $r$.}
  \vspace{-4 mm}
  \label{fig:0104}
\end{figure}

Direct prefix extraction retains coordinates whose positions need not reflect their reconstruction importance. Fig.~\ref{fig:0104} contrasts this choice with a best rank-constrained approximation. SP+ constructs a new shared adapter representation whose prefixes solve
\begin{equation}
M_{r_i}^{+}
\in\operatorname*{arg\,min}_{\operatorname{rank}(Z)\leq r_i}
\left\|M^g-Z\right\|_F^2.
\label{eq:spplus-objective}
\end{equation}

\begin{figure}[t]
  \centering
  \includegraphics[width=0.95\linewidth]{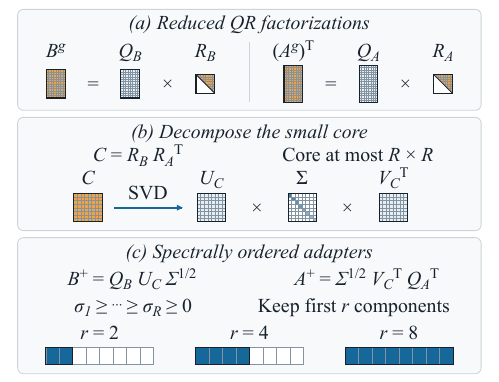}
  \caption{Rank-specific adapter extraction using established QR/core-SVD decomposition~\cite{rblaMethodBallaniKressner2016}. (a) Reduced QR factorizations produce orthonormal bases and small triangular matrices. (b) SVD is applied to the core $C=R_BR_A^{\mathsf T}$, whose dimensions are at most $R\times R$. (c) The resulting adapters preserve $B^gA^g$ and provide ordered prefixes for client-specific ranks without forming the dense product.}
  \vspace{-4 mm}
  \label{fig:0102}
\end{figure}

Because $s_s$ is positive, $\Delta W^g$ and $M^g$ have the same singular directions. SP+ obtains these directions using established QR/core-SVD decomposition~\cite{rblaMethodBallaniKressner2016}, as illustrated in Fig.~\ref{fig:0102}, without forming either dense matrix. The server first computes reduced QR factorizations:
\begin{equation}
B^g=Q_BR_B,\qquad
(A^g)^{\mathsf T}=Q_AR_A,
\label{eq:spplus-qr}
\end{equation}
where $Q_B\in\mathbb{R}^{d_{\mathrm{out}}\times R}$ and $Q_A\in\mathbb{R}^{d_{\mathrm{in}}\times R}$ have orthonormal columns, and $R_B,R_A\in\mathbb{R}^{R\times R}$. It then forms the core matrix
\begin{equation}
C=R_BR_A^{\mathsf T}\in\mathbb{R}^{R\times R}
\label{eq:spplus-core}
\end{equation}
and computes its SVD:
\begin{equation}
\begin{gathered}
C=U_C\Sigma V_C^{\mathsf T},\\
\Sigma=\operatorname{diag}(\sigma_1,\ldots,\sigma_R).
\end{gathered}
\label{eq:spplus-core-svd}
\end{equation}
Here, $U_C,\Sigma,V_C\in\mathbb{R}^{R\times R}$, and $U_C,V_C$ are orthogonal. The singular values are ordered as $\sigma_1\geq\cdots\geq\sigma_R\geq0$, including zeros when $M^g$ is rank deficient. This yields the implicit decomposition
\begin{equation}
M^g=(Q_BU_C)\Sigma(Q_AV_C)^{\mathsf T}.
\label{eq:spplus-implicit-svd}
\end{equation}
Thus, $\sigma_j$ are singular values of $M^g$, and $s_s\sigma_j$ are those of $\Delta W^g$. The server represents the ordered components using balanced factors:
\begin{equation}
\begin{aligned}
B^+&=Q_BU_C\Sigma^{1/2},\\
A^+&=\Sigma^{1/2}V_C^{\mathsf T}Q_A^{\mathsf T}.
\end{aligned}
\label{eq:spplus-factors}
\end{equation}
Their product preserves the complete effective update in exact arithmetic:
\begin{equation}
s_sB^+A^+=s_sB^gA^g=\Delta W^g.
\label{eq:spplus-preservation}
\end{equation}
The square root of $\Sigma$ assigns equal singular-value scaling to the two factors. This balancing preserves their width and product. Client-scale conversion below preserves the intended effective update when the receiving module uses a different coefficient.

For a client with rank $r_i$, the server selects the leading columns and rows of the balanced factors:
\begin{equation}
B^+_{:,1:r_i},\qquad A^+_{1:r_i,:}.
\label{eq:spplus-client-factors}
\end{equation}
Let $u_j$ and $v_j$ denote the columns of $Q_BU_C$ and $Q_AV_C$, respectively. The selected factors represent the target effective update
\begin{equation}
s_sM_{r_i}^{+}
=s_sB^+_{:,1:r_i}A^+_{1:r_i,:}
=s_s\sum_{j=1}^{r_i}\sigma_j u_jv_j^{\mathsf T}.
\label{eq:spplus-client-update}
\end{equation}
The unscaled product $M_{r_i}^+$ solves Eq.~\eqref{eq:spplus-objective} and also minimizes the unsquared Frobenius error used in Section~\ref{sec:problem-statement}. Its effective update has squared approximation error
\begin{equation}
\left\|\Delta W^g-s_sM_{r_i}^{+}\right\|_F^2
=s_s^2\sum_{j=r_i+1}^{R}\sigma_j^2.
\label{eq:spplus-truncation-error}
\end{equation}
Therefore, a single decomposition serves all client ranks, with larger prefixes retaining the components of smaller prefixes. The reconstruction error is zero whenever $r_i\geq\operatorname{rank}(M^g)$, including $r_i=R$. At each supported rank, the client incurs only the unavoidable rank-constrained error for the same fixed aggregate. Its effect on task accuracy is evaluated experimentally. The selected factors remain expressed under the server's scaling convention.

\subsubsection{Client-scale conversion and broadcast}
Client $i$ applies its own coefficient $s_i$. Directly loading the selected server factors would give
\begin{equation}
s_iB^+_{:,1:r_i}A^+_{1:r_i,:}=s_iM_{r_i}^{+}.
\label{eq:rank-scale-mismatch}
\end{equation}

For example, $s_i=4s_s$ multiplies the retained update by four. The server prevents this amplitude mismatch by converting the extracted pair before transmission:
\begin{equation}
\begin{gathered}
d_i=\sqrt{\frac{s_s}{s_i}},\\
B_i^{\mathrm{next}}=d_iB^+_{:,1:r_i},\qquad
A_i^{\mathrm{next}}=d_iA^+_{1:r_i,:}.
\end{gathered}
\label{eq:rank-scale-broadcast}
\end{equation}

Therefore, the receiving client reproduces the intended rank-specific effective update:
\begin{equation}
s_iB_i^{\mathrm{next}}A_i^{\mathrm{next}}
=s_id_i^2B^+_{:,1:r_i}A^+_{1:r_i,:}
=s_sM_{r_i}^{+}.
\label{eq:rank-scale-broadcast-preservation}
\end{equation}

As in the upload, the square-root coefficient follows from applying the same positive multiplier to both factors. With a common numerator and $s=\alpha/r$, it reduces to $d_i=\sqrt{r_i/r_s}$. When $s_i=s_s$, both conversions are identities. Scalar multiplication commutes with prefix selection, so the conversion can be performed before or after selecting the client's prefix. For a fixed aggregate, the retained spectral components are orthogonal to the discarded components in the Frobenius inner product. Therefore, omitting the downlink conversion produces the error decomposition
\begin{equation}
\begin{aligned}
&\left\|s_iM_{r_i}^{+}-\Delta W^g\right\|_F^2\\
&\quad=
\left\|s_sM_{r_i}^{+}-\Delta W^g\right\|_F^2\\
&\qquad+
\left(\frac{s_i}{s_s}-1\right)^2
\left\|s_sM_{r_i}^{+}\right\|_F^2.
\end{aligned}
\label{eq:rank-scale-error-decomposition}
\end{equation}

The first term is the unavoidable rank-constrained error in Eq.~\eqref{eq:spplus-truncation-error}. Conversion removes the second term, which results from an incorrect amplitude of the retained components. It does not recover discarded components. Together with spectral extraction, it minimizes Frobenius reconstruction error for the fixed aggregate within each client's rank budget, without requiring client data. This avoids additional reconstruction error without increasing the prescribed adapter size. A smaller Frobenius error also tightens an upper bound on the perturbation to the layer output for a given input.

\section{Experiments}
\label{sec:experiments}

We evaluate whether rank-specific LoRA adapter extraction can preserve useful updates within prescribed client budgets while keeping server processing costs low. The experiments address three questions. First, how does RBLA+ affect task accuracy when clients use different ranks? Second, does compact extraction preserve the fixed aggregate, and how does this relate to subsequent training? Third, what computation, memory, and adapter payload are required to prepare client updates? We use federated training simulations for the first two questions and separate CPU measurements and payload analysis for the third.

\subsection{Experimental Setup}
\label{subsec:experimental-setup}

\subsubsection{Tasks, Data partitions, and Methods} We evaluate RBLA+ on seven image classification datasets: MNIST~\cite{lecun1998mnist}, FMNIST~\cite{xiao2017fashionmnist}, KMNIST~\cite{clanuwat2018kmnist}, QMNIST~\cite{yadav2019qmnist}, CIFAR-10~\cite{krizhevsky2009cifar}, CINIC-10~\cite{darlow2018cinic10}, and SVHN~\cite{netzer2011svhn}. We also evaluate six language classification tasks: AG News~\cite{zhang2015character}, IMDB~\cite{maas2011learning}, and the GLUE tasks MNLI, QQP, RTE, and SST-2~\cite{wang2018glue}. Each dataset is partitioned among ten clients using a Dirichlet distribution~\cite{hsu2019measuring}. We denote the concentration parameter by $\alpha_{\mathrm D}$ to distinguish it from the LoRA scaling parameter. Lower values produce more uneven class allocations across clients in expectation. The main experiments use $\alpha_{\mathrm D}=0.4$ for all 13 datasets under two rank assignments, yielding 26 evaluation settings. We compare RBLA+ with six baselines: RBLA~\cite{chen2024rbla}, FlexLoRA~\cite{bai2024federated}, Zero Padding~\cite{cho2024heterogeneous}, Replication Padding~\cite{byun2024towards}, FLoRIST~\cite{ramesh2025florist}, and FLoRA~\cite{wang2024flora}.

\subsubsection{Client ranks} We use prescribed ranks to represent different client adapter budgets. The \emph{correct} assignment gives larger rank ratios to clients with more samples, whereas the \emph{mismatched} assignment reverses this ordering. These names describe the relation between data volume and adapter size. Let $n_i$ be client $i$'s data volume, with $n_{\min}=\min(n_i)$ and $n_{\max}=\max(n_i)$. For the color-image and language runs, the rank ratios follow
\begin{equation}
\begin{aligned}
\rho_i^{\mathrm{correct}}&=0.25+0.75\frac{n_i-n_{\min}}{n_{\max}-n_{\min}},\\
\rho_i^{\mathrm{mismatched}}&=1.25-\rho_i^{\mathrm{correct}},
\end{aligned}
\label{eq:experiment-rank-assignment}
\end{equation}
with $n_{\max}>n_{\min}$ in these partitions. Rank assignments remain fixed during training.

\subsubsection{Models and training} The experiment configurations specify ten participating clients per round and set the training-round parameter to 100. The MNIST-family model is a multilayer perceptron with layer widths $784$, $200$, $200$, and $10$. The MLP use relatively high adapter ranks to evaluate aggregation and redistribution under a large total rank budget. Since the uncompressed stacked representation has width $S=\sum_i r_i$, this configuration exposes the representation-size pressure that can also arise with many participating clients. The classification-layer rank remains capped by its matrix dimensions. CIFAR-10 and CINIC-10 use a low-rank-parameterized CNN, SVHN uses VGG9, and all language tasks use Tiny-BERT. For the MNIST-family, CIFAR-10, CINIC-10 and SVHN tasks, the LoRA adapters are trained from scratch, without a pretrained weight term $W_0$. These settings examine aggregation and rank-specific redistribution when learning relies directly on low-rank factors, extending the evaluation to low-rank federated training without pretrained initialization. In contrast, Tiny-BERT NLP tasks are initialized with pretrained weights and adapted using LoRA to evaluate the methods in pretrained-model fine-tuning settings. CNN and VGG9 ranks are $\max(1,\operatorname{round}(b\rho_i))$, with convolutional base ranks $b=64$ and $8$, respectively, and base rank $16$ for adapted fully connected layers. Tiny-BERT uses $\max(1,\operatorname{round}(8\rho_i))$. Training experiments were conducted on an AWS EC2 g5.2xlarge instance with one NVIDIA A10G GPU. Server adapter-extraction costs were measured separately on a local device equipped with an AMD Ryzen 9 9950X CPU. For MNIST-family experiment seed were set to 1,2,3 and rest experiment seeds are set to fixed 42.

\subsubsection{Evaluation} Tables~\ref{tab:correct-rank-assignment} and~\ref{tab:mismatched-rank-assignment} report highest global model accuracy during training. Entries with $\pm$ report the mean and standard deviation of peak from each three runs.

\subsection{Accuracy under Prescribed Client LoRA Ranks}
\label{subsec:experiment-accuracy}

Tables~\ref{tab:correct-rank-assignment} and~\ref{tab:mismatched-rank-assignment} examine task accuracy under the prescribed rank assignments at $\alpha_{\mathrm D}=0.4$. Each comparison uses the same assignment across methods. The header also reports analytical downlink payload, allowing accuracy to be read alongside the size of the adapters sent to clients. RBLA+ exceeds RBLA in 24 of the 26 conditions and attains the highest reported value among the 7 methods in 16 conditions: 9 under correct assignment and 7 under mismatched assignment These counts describe the reported means and point estimates.

\begin{table*}[t]
\centering
\caption{Highest test accuracy (\%) under the correct rank assignment with Dirichlet distribution $\alpha_{\mathrm D}=0.4$.}
\label{tab:correct-rank-assignment}
\footnotesize
\setlength{\tabcolsep}{3pt}
\renewcommand{\arraystretch}{1.08}
\begin{tabular*}{\textwidth}{@{\extracolsep{\fill}}lrrrrrrr@{}}
\hline
\shortstack[l]{\strut Dataset / Method\\
               \strut Communication overhead}
& \shortstack{\strut RBLA\\\strut($1\times$)}
& \shortstack{\strut RBLA+\\\strut($1\times$)}
& \shortstack{\strut FlexLoRA\\\strut($1\times$)}
& \shortstack{\strut Zero Padding\\\strut($1\times$)}
& \shortstack{\strut Rep. Padding\\\strut($1\times$)}
& \shortstack{\strut FLoRIST\\\strut($1\times$)}
& \shortstack{\strut FLoRA\\\strut($10\times$)} \\
\hline
MNIST
& $79.54\pm2.91$
& $\mathbf{82.01\pm2.27}$
& $42.76\pm22.57$
& $79.18\pm3.36$
& $78.90\pm3.32$
& $35.87\pm2.98$
& $44.32\pm8.32$ \\
FMNIST
& $62.53\pm1.48$
& $\mathbf{65.89\pm2.98}$
& $40.06\pm11.18$
& $62.70\pm3.40$
& $62.29\pm1.30$
& $35.37\pm10.30$
& $48.72\pm5.99$ \\
KMNIST
& $41.59\pm4.81$
& $\mathbf{43.85\pm2.78}$
& $16.01\pm1.19$
& $41.21\pm5.03$
& $41.71\pm4.98$
& $22.90\pm1.85$
& $19.05\pm2.35$ \\
QMNIST
& $79.29\pm3.27$
& $\mathbf{81.63\pm2.19}$
& $42.49\pm22.45$
& $78.81\pm3.76$
& $78.65\pm3.41$
& $36.10\pm3.62$
& $44.01\pm8.01$ \\
\hline
CIFAR-10
& 38.33 & \textbf{79.12} & 78.84 & 78.39
& 45.61 & 72.18 & 77.25 \\
CINIC-10
& 31.54 & \textbf{69.65} & 69.33 & 68.06
& 26.31 & 60.10 & 68.51 \\
SVHN
& 72.39 & \textbf{92.98} & 92.70 & 91.89
& 73.05 & 90.96 & 92.23 \\
\hline
AG News
& 90.68 & 91.33 & 90.75 & 89.50
& 90.50 & 89.13 & \textbf{91.39} \\
IMDB
& 84.15 & \textbf{84.32} & 81.89 & 81.22
& 84.16 & 81.30 & 83.65 \\
MNLI
& 50.25 & 65.16 & 61.67 & 62.34
& 47.52 & 63.16 & \textbf{65.28} \\
QQP
& 78.50 & 82.48 & 81.20 & 77.94
& 77.14 & 78.84 & \textbf{83.47} \\
RTE
& 55.23 & \textbf{57.04} & 55.60 & 55.96
& 55.60 & 55.96 & 53.07 \\
SST-2
& 78.44 & 80.62 & 79.70 & 80.62
& 77.98 & 79.82 & \textbf{81.31} \\
\hline
\end{tabular*}
\end{table*}

\begin{table*}[t]
\centering
\caption{Highest test accuracy (\%) under the mismatched rank assignment with Dirichlet distribution $\alpha_{\mathrm D}=0.4$.}
\label{tab:mismatched-rank-assignment}
\footnotesize
\setlength{\tabcolsep}{3pt}
\renewcommand{\arraystretch}{1.08}
\begin{tabular*}{\textwidth}{@{\extracolsep{\fill}}lrrrrrrr@{}}
\hline
\shortstack[l]{\strut Dataset / Method\\
               \strut Communication overhead}
& \shortstack{\strut RBLA\\\strut($1\times$)}
& \shortstack{\strut RBLA+\\\strut($1\times$)}
& \shortstack{\strut FlexLoRA\\\strut($1\times$)}
& \shortstack{\strut Zero Padding\\\strut($1\times$)}
& \shortstack{\strut Rep. Padding\\\strut($1\times$)}
& \shortstack{\strut FLoRIST\\\strut($1\times$)}
& \shortstack{\strut FLoRA\\\strut($10\times$)} \\
\hline
MNIST
& $75.00\pm1.95$
& $\mathbf{77.58\pm3.52}$
& $39.03\pm19.25$
& $73.90\pm2.51$
& $74.21\pm2.59$
& $35.95\pm8.34$
& $46.23\pm8.96$ \\
FMNIST
& $\mathbf{58.91\pm7.41}$
& $58.26\pm8.80$
& $33.38\pm3.19$
& $57.54\pm8.55$
& $58.49\pm7.71$
& $33.40\pm2.28$
& $48.50\pm6.57$ \\
KMNIST
& $37.44\pm1.50$
& $\mathbf{39.94\pm1.71}$
& $22.00\pm4.63$
& $36.93\pm1.26$
& $37.23\pm1.05$
& $22.78\pm0.59$
& $19.95\pm1.66$ \\
QMNIST
& $74.69\pm2.55$
& $\mathbf{77.37\pm3.03}$
& $38.32\pm18.10$
& $73.74\pm2.70$
& $74.10\pm2.81$
& $33.09\pm5.99$
& $46.03\pm8.34$ \\
\hline
CIFAR-10
& 57.63 & \textbf{78.95} & 78.53 & 78.22
& 41.08 & 71.27 & 78.29 \\
CINIC-10
& 42.10 & \textbf{68.89} & 68.33 & 68.46
& 34.03 & 61.25 & 68.61 \\
SVHN
& 90.35 & \textbf{93.20} & 91.99 & 90.44
& 90.97 & 90.76 & 91.84 \\
\hline
AG News
& 90.28 & 91.68 & 89.32 & 85.36
& 88.21 & 89.34 & \textbf{91.86} \\
IMDB
& 83.10 & \textbf{83.64} & 80.97 & 81.84
& 83.33 & 81.91 & 82.38 \\
MNLI
& 57.80 & 64.37 & 46.32 & 43.53
& 56.51 & 63.08 & \textbf{65.49} \\
QQP
& 73.78 & 81.02 & 72.57 & 63.40
& 73.22 & 78.75 & \textbf{82.39} \\
RTE
& 53.43 & 55.60 & 54.51 & 54.87
& 53.79 & \textbf{58.48} & 49.82 \\
SST-2
& \textbf{81.08} & 80.50 & 80.16 & 79.36
& 80.28 & 80.05 & \textbf{81.08} \\
\hline
\end{tabular*}
\vspace{-4mm}
\end{table*}

\subsubsection{Vision tasks}

Under correct assignment, RBLA+ achieves the highest mean accuracy on all four MNIST-family datasets, exceeding RBLA by 2.26 to 3.36 percentage points (pp). Under mismatched assignment, it has the highest reported mean on MNIST, KMNIST, and QMNIST, while its FMNIST mean is 0.65 pp below RBLA. The largest gains over RBLA occur on CIFAR-10 and CINIC-10: 40.79 and 38.11 pp under correct assignment, and 21.32 and 26.79 pp under mismatched assignment. On these two datasets, RBLA+ also exceeds the strongest alternative by 0.28 to 0.42 pp. On SVHN, it reaches 92.98\% and 93.20\% under correct and mismatched assignment, respectively, improving over the strongest alternative by 0.28 and 1.21 pp. Across all 14 vision conditions, RBLA+ reports higher accuracy than FLoRA while requiring substantially less down-link adapter communication.

\subsubsection{Language tasks}

RBLA+ achieves the highest reported accuracy on IMDB under both assignments and on RTE under the correct assignment. On AG News, MNLI, QQP, and SST-2, it trails FLoRA by 0.06 to 1.37 pp across the two assignments, while improving over RBLA on MNLI and QQP. The gains over RBLA are not uniform. On SST-2 under mismatched assignment, RBLA+ reaches 80.50\%, below RBLA and FLoRA at 81.08\%. Under correct assignment, its 80.62\% ties Zero Padding but remains below FLoRA at 81.31\%. These results show that RBLA+ achieves accuracy close to FLoRA on four language tasks and higher accuracy on the other two, with a substantially smaller broadcast payload. With ten clients, its analytical adapter-only downlink payload, summed over all clients per round, is only one tenth of FLoRA's, corresponding to a 90\% reduction. Section~\ref{subsec:experiment-cost} details this payload comparison. Small differences between single-run accuracy results remain descriptive comparisons.

\subsubsection{Random client participation}

\begin{table}[t]
\centering
\caption{Peak global-model accuracy (\%) with 100 clients
and 10 randomly selected participants per round.}
\label{tab:random-client-selection}
\small
\renewcommand{\arraystretch}{1.08}
\begin{tabular*}{\columnwidth}{@{\extracolsep{\fill}}lrr@{}}
\hline
Method & CIFAR-10 & MNLI \\
\hline
RBLA         & 75.17 & 56.17 \\
RBLA+        & \textbf{75.21} & \textbf{61.21} \\
FlexLoRA     & 73.94 & 52.93 \\
Zero Padding & 72.34 & 56.43 \\
Rep. Padding & 74.54 & 56.40 \\
FLoRIST      & 56.84 & 56.79 \\
FLoRA        & 64.07 & 57.91 \\
\hline
\end{tabular*}
\vspace{-4mm}
\end{table}

Table~\ref{tab:random-client-selection} evaluates partial participation with 100 clients, of which 10 are randomly selected in each communication round. The training sets are repartitioned using Dirichlet $\alpha_{\mathrm D}=0.4$, and client ranks are reassigned using the same correct rank assignment rule. All methods share the same partition and participant sequence with seed 42; the remaining training hyper-parameters are unchanged. RBLA+ achieves the highest peak global-model accuracy on both datasets: 75.21\% on CIFAR-10, closely matching RBLA's 75.17\%, and 61.21\% on MNLI, exceeding RBLA and the strongest competing baseline, FLoRA, by 5.04 and 3.30 pp, respectively. Compared with the full-participation results, the CIFAR-10 results also show substantial changes in several baselines. RBLA improves from 38.33\% to 75.17\%, and Rep.~Padding from 45.61\% to 74.54\%, whereas Zero Padding decreases from 78.39\% to 72.34\%. To interpret these changes, we examine two consequences of the new configuration. Repartitioning the same dataset among more clients reduces local optimization steps on average at fixed local epochs and batch size, potentially limiting client drift. Random participation also changes the maximum adapter width available in each round, which interacts with the broadcast policy described below. Although only 10 clients are trained per communication round, all 100 stored client adapter states are updated from each aggregation. When an aggregated weight is narrower than a client's weight width, we extend both factors with zeros. This provides a deterministic extension that preserves the received product, $B_{\mathrm{pad}}A_{\mathrm{pad}}=BA$, without additional tail initialization or retention of stale client parameters. For a given layer, let $w_t^{\max}=\max_{i\in\mathcal S_t}w_i$, where $\mathcal S_t$ is the participating set and $w_i$ is the actual factor width. For RBLA, Rep.~Padding, and Zero Padding, updating all client states sets both factors to zero in slots beyond $w_t^{\max}$. Since $a_j=b_j=0$ implies $\nabla_{a_j}\mathcal L=\nabla_{b_j}\mathcal L=0$ for the backward propagation $\mathcal L$ is the local training loss. These slots remain inactive under local training with reset optimizer states and no tail re-initialization. Consequently, their potentially active slot indices satisfy $j\le h_t$, where $h_t=\min_{\tau\le t}w_\tau^{\max}$. Table~\ref{tab:random-active-slot-bounds} reports these structural upper bounds, derived from the participant sequence and broadcast policy. Thus, the contraction of potentially active slots occurs in both the CNN and language-model settings.

\begin{table}[t]
\centering
\caption{Shared upper bounds on potentially nonzero factor slots for RBLA, Rep.~Padding, and Zero Padding under random participation. Bold entries indicate decreases.}
\label{tab:random-active-slot-bounds}
\small
\setlength{\tabcolsep}{3pt}
\renewcommand{\arraystretch}{1.15}
\begin{tabular*}{\columnwidth}{@{\extracolsep{\fill}}lrrrrrrrr@{}}
\hline
Round
& 0 & 4 & 5 & 7 & 24 & 25 & 35 & 55 \\
\hline
CIFAR-10 Conv.
& 53 & \textbf{51} & \textbf{44} & \textbf{41}
& \textbf{35} & 35 & 35 & \textbf{34} \\
CIFAR-10 FC
& 13 & 13 & \textbf{11} & \textbf{10}
& \textbf{9} & 9 & 9 & 9 \\
MNLI
& 5 & 5 & 5 & 5
& 5 & \textbf{4} & \textbf{3} & 3 \\
\hline
\end{tabular*}
\vspace{-4mm}
\end{table}

A possible explanation for the contrasting baseline changes is that shorter local optimization limits client drift, while tail suppression reduces unfavorable high-index contributions in RBLA and Rep.~Padding. Zero Padding already dilutes low-support contributions during aggregation: for identical uploaded factors, normalized aggregation weights, and scaling conventions, $M_j^{\mathrm{Zero}}=c_j^2M_j^{\mathrm{RBLA}}$, where $M_j=b_ja_j^\top$ and $c_j$ is the total normalized weight of participating clients containing slot $j$. Further restricting these slots may therefore remove useful capacity. 

\subsubsection{Reconstruction of global factors} 
\begin{figure}[t]
\centering
\begin{subfigure}[t]{0.48\linewidth}
\centering
\includegraphics[width=\linewidth]{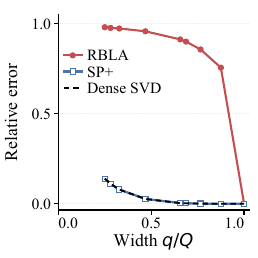}
\caption{CIFAR-10}
\label{fig:checkpoint_reconstruction_cifar10}
\end{subfigure}\hfill
\begin{subfigure}[t]{0.48\linewidth}
\centering
\includegraphics[width=\linewidth]{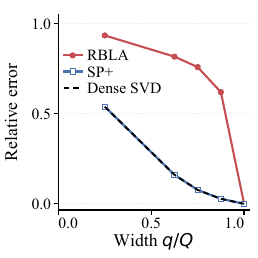}
\caption{MNLI}
\label{fig:checkpoint_reconstruction_mnli}
\end{subfigure}
\caption{Reconstruction of trained RBLA aggregates at round 100. $Q$ denotes the aggregate factor width of the selected layer, and $q$ denotes the prefix width. SP+ matches the dense SVD reference within numerical precision across the tested adapter widths. Raw prefixes incur additional reconstruction error. }
\label{fig:checkpoint-reconstruction}\vspace{-4mm}
\end{figure}

Fig.~\ref{fig:checkpoint-reconstruction} uses the aggregated LoRA adapters collected at communication round 100 during RBLA training as inputs to different adapter extraction methods. No additional gauge transformation is applied. For each task, we select the middle adapted layer in structural order: \texttt{conv2b} on CIFAR-10 and the first encoder block's feed-forward output projection on MNLI. Saved adapters are converted to FP64 for reconstruction and error evaluation. At $q/Q=0.25$, the CIFAR-10 relative error decreases from 0.980 for the raw prefix to 0.140 for SP+. The MNLI error decreases from 0.934 to 0.539. SP+ overlaps the dense optimum across the tested widths. For these two layers, SP+ retains more of the same trained aggregate without increasing the client's adapter width or analytical payload.

\subsubsection{Rank needed for a target error}
\begin{figure}[t]
\centering
\begin{subfigure}[t]{0.48\linewidth}
\centering
\includegraphics[width=\linewidth]{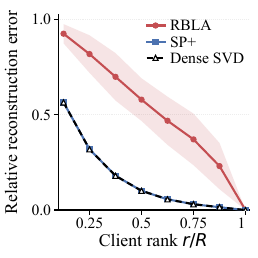}
\caption{Error at a given rank.}
\label{fig:broadcast_reconstruction_error}
\end{subfigure}\hfill
\begin{subfigure}[t]{0.48\linewidth}
\centering
\includegraphics[width=\linewidth]{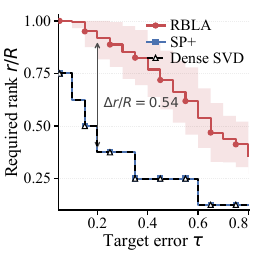}
\caption{Rank at a target error.}
\label{fig:matched_error_rank_requirement}
\end{subfigure}
\caption{Reconstruction of a fixed synthetic rank-$8$ update. Curves show means over 20 orthogonal rotations. Bands show one sample standard deviation. SP+ overlaps the dense SVD reference.}
\label{fig:rank_error}\vspace{-4mm}
\end{figure}

Fig.~\ref{fig:rank_error} asks the complementary question of how much rank is needed to meet a reconstruction threshold. It uses a fixed $128\times96$ synthetic update with rank $R=8$ and singular values $\sigma_j=\exp[-4(j-1)/7]$. We generate 20 orthogonal rotations using seeds 42 to 61. For each rotation and threshold $\tau\in\{0.05,0.10,\ldots,0.80\}$, we select the smallest integer rank in $\{1,\ldots,8\}$ with $E_r/\|M^g\|_F\le\tau$, without interpolation. At $\tau=0.2$, SP+ and dense SVD require $r/R=0.375$, whereas the mean requirement for RBLA prefixes is 0.91875. This fixed-matrix result shows that an ordered spectral representation can meet the same reconstruction target with fewer adapter entries.

\subsection{Extraction and the Training Trajectory}
\label{subsec:experiment-broadcast-analysis}

\subsubsection{Changing the evaluated extraction} 
\begin{figure}[t]
\centering
\begin{subfigure}[t]{0.48\linewidth}
\centering
\includegraphics[width=\linewidth]{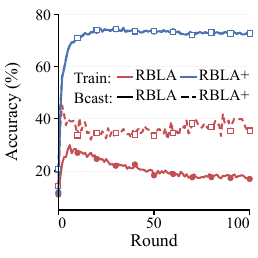}
\caption{CIFAR-10}
\label{fig:training_broadcast_accuracy_cifar10}
\end{subfigure}\hfill
\begin{subfigure}[t]{0.48\linewidth}
\centering
\includegraphics[width=\linewidth]{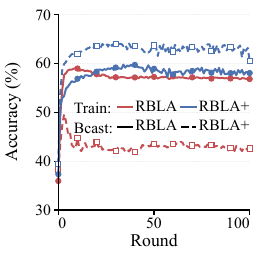}
\caption{MNLI}
\label{fig:training_broadcast_accuracy_mnli}
\end{subfigure}
\caption{Training-path and broadcast ablation. This experiment assesses broadcast quality through the accuracy of the client models. At each evaluated round, both broadcasts use the same aggregated model within each training path to construct models at the ten clients' assigned ranks. Curves report their mean accuracy on a common test set before local training. Color identifies the training path (Train), and line style identifies the evaluated broadcast (Bcast). Each path uses a single training seed. Alternate-broadcast evaluations do not affect subsequent training.}
\label{fig:training-broadcast-accuracy}\vspace{-4mm}
\end{figure}

A better approximation of an aggregate need not yield a better client predictor. Fig.~\ref{fig:training-broadcast-accuracy} uses seed 42 with correctly assigned client ranks. At each round, both broadcasts are evaluated from the same aggregated LoRA factors within each training path before next round training, without affecting subsequent training. Curves show mean local model accuracy across ten clients. At round 100 on CIFAR-10, switching from RBLA to RBLA+ broadcast along the RBLA path raises accuracy from 16.97\% to 35.49\%, still below the 72.73\% achieved by the RBLA+ path with its own broadcast. On MNLI, the same replacement instead lowers accuracy from 56.81\% to 42.60\%.

A possible explanation for the MNLI degradation is training under prefix-based broadcasts enhances task-relevant behavior that can't preserved by a one-shot spectral extraction. Frobenius-optimal reconstruction preserves the aggregated LoRA factors in parameter space, but need not preserve the original prefixes' predictions or improve their accuracy. Repeated spectral broadcasting, in contrast, also changes subsequent local optimization and the aggregates produced in later rounds. The degradation under a one-shot replacement is therefore compatible with the stronger performance of the full RBLA+ training path. On CIFAR-10, the improvement is consistent with spectral reconstruction reducing the additional error introduced by prefix extraction. This may lessen perturbations to intermediate features and improve prediction, but a one-shot replacement cannot recover the benefits accumulated through repeated spectral broadcasting during training.

\subsubsection{Prefix quality along training}

\begin{figure}[t]
\centering
\begin{subfigure}[t]{0.48\linewidth}
\centering
\includegraphics[width=\linewidth]{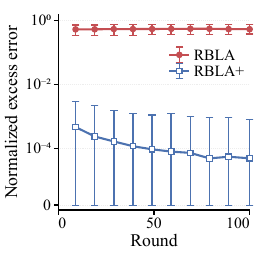}
\caption{CIFAR-10}
\label{fig:prefix_reconstruction_quality_cifar10}
\end{subfigure}\hfill
\begin{subfigure}[t]{0.48\linewidth}
\centering
\includegraphics[width=\linewidth]{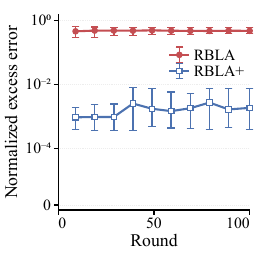}
\caption{MNLI}
\label{fig:prefix_reconstruction_quality_mnli}
\end{subfigure}
\caption{Raw-prefix quality along the RBLA and RBLA+ training paths, measured by the normalized excess squared error $\delta_q$ at ten checkpoints. Markers and error bars indicate medians and interquartile ranges across layer/width pairs with $q<Q$, respectively. Each prefix is evaluated against the rank-$q$ optimum of its own reference aggregate. The vertical axis uses a symmetric logarithmic scale.}
\label{fig:prefix-reconstruction-quality}\vspace{-4mm}
\end{figure}

Fig.~\ref{fig:prefix-reconstruction-quality} examines the raw prefixes of the pre-spectral aggregates produced along the RBLA and RBLA+ training trajectories. For each non-zero aggregated global factor, we measure the normalized excess reconstruction error as:
\begin{equation}
\delta_q=
\frac{\|M^g-M_q^{\mathrm{pre}}\|_F^2-(E_q^\star)^2}
     {\|M^g\|_F^2},
\label{eq:experiment-excess-squared-error}
\end{equation}
where $E_q^\star$ is the minimum rank-$q$ reconstruction error of the same aggregated update, as defined in Eq.~\eqref{eq:optimal-rank-r-error}. We evaluate ten checkpoints at rounds 9, 19, 29, 39, 49, 59, 69, 79, 89, and 100. For each training trajectory and checkpoint, each marker reports the median over all layers and distinct requested widths satisfying $q<Q$. Across both datasets, the RBLA+ trajectory yields substantially lower median normalized excess errors than the RBLA trajectory at every sampled checkpoint. Thus, the improvement extends beyond an individual layer or checkpoint: the raw prefixes of aggregates produced along the RBLA+ trajectory remain closer to their respective rank-$q$ optima throughout the sampled training stages. At round 100, the median $\delta_q$ is 0.523 along the CIFAR-10 RBLA training path and $4.84\times10^{-5}$ along the RBLA+ path. The corresponding MNLI values are 0.466 and $1.83\times10^{-3}$. At all ten inspected checkpoints, the raw prefixes from the RBLA+ training path exhibit lower median normalized excess squared error. These results extend the layer-level reconstruction analysis across multiple stages of training. Together with the accuracy results in Fig.~\ref{fig:training-broadcast-accuracy}, they characterize both the reconstruction quality and predictive performance of the two training paths.

\subsubsection{Scale-conversion comparison}
Table~\ref{tab:rbla-rank-scale-accuracy} compares RBLA+ runs with and without scale conversion at $\alpha_{\mathrm D}=0.4$. Runs with conversion report higher peak global-model accuracy in 13 of the 18 comparisons, including ten of the 12 language comparisons. A possible explanation for the task-dependent decreases under mismatched rank assignment correlated with the interaction between rank-dependent scaling and local data coverage. In this scenario, higher-rank clients receive fewer training samples, so their additional components may emphasize local features that are less representative of the global distribution. When client scalings differ, omitting conversion introduces extra amplification at lower-rank clients, which hold more data under the mismatched assignment. Removing this amplification may increase the relative influence of higher-rank, data-poor clients along the resulting training trajectory. Their locally specialized updates could then adversely affect global generalization on some tasks.

\begin{table}[t]
  \centering
   \caption{Peak global model accuracy (\%) of RBLA+ with and without explicit scale conversion. Differences are computed before rounding.}
  \label{tab:rbla-rank-scale-accuracy}
  \footnotesize
  \setlength{\tabcolsep}{3pt}
  \begin{tabular*}{\columnwidth}{@{\extracolsep{\fill}}lrrr@{}}
    \toprule
    Dataset & Without conversion & With conversion & $\Delta$ (pp) \\
    \midrule
    \multicolumn{4}{c}{\textit{Correct rank assignment}} \\
    \midrule
    CIFAR-10 & 78.30 & \textbf{79.12} & +0.82 \\
    CINIC-10 & 69.13 & \textbf{69.65} & +0.52 \\
    SVHN & 92.90 & \textbf{92.98} & +0.08 \\
    AG News & 90.99 & \textbf{91.33} & +0.34 \\
    IMDB & 82.25 & \textbf{84.32} & +2.07 \\
    MNLI & 64.51 & \textbf{65.16} & +0.64 \\
    QQP & 81.66 & \textbf{82.48} & +0.82 \\
    RTE & 54.51 & \textbf{57.04} & +2.53 \\
    SST-2 & 79.36 & \textbf{80.62} & +1.26 \\
    \midrule
    \multicolumn{4}{c}{\textit{Mismatched rank assignment}} \\
    \midrule
    CIFAR-10 & \textbf{79.21} & 78.95 & -0.26 \\
    CINIC-10 & \textbf{69.25} & 68.89 & -0.36 \\
    SVHN & \textbf{93.45} & 93.20 & -0.25 \\
    AG News & 90.50 & \textbf{91.68} & +1.18 \\
    IMDB & 81.41 & \textbf{83.64} & +2.23 \\
    MNLI & 60.02 & \textbf{64.37} & +4.35 \\
    QQP & 79.81 & \textbf{81.02} & +1.21 \\
    RTE & \textbf{55.96} & 55.60 & -0.36 \\
    SST-2 & \textbf{81.08} & 80.50 & -0.57 \\
    \bottomrule
  \end{tabular*}
  \vspace{-4mm}
\end{table}

\subsection{Server Extraction Cost and Adapter Communication}
\label{subsec:experiment-cost}

\begin{table}[t]
\centering
\caption{CPU adapter-extraction and output-preparation costs for RBLA, RBLA+, and alternative decomposition methods. Counts beside model names give the selected adapter parameters for the maximum-rank client, in millions.}
\label{tab:accuracy-extraction-cost}
\small
\setlength{\tabcolsep}{3pt}
\renewcommand{\arraystretch}{1.08}
\begin{tabular*}{\columnwidth}
{@{\extracolsep{\fill}}lrr@{}}
\hline
Method & Time (ms) & $\Delta$RSS (MiB) \\
\hline
\multicolumn{3}{c}{\textit{MLP: 0.224M}} \\
RBLA                 & 0.4   & 5.5  \\
RBLA+ (QR-SVD)        & 5.0   & 13.5 \\
Dense full SVD        & 6.9   & 13.6 \\
Dense partial SVD     & 88.3  & 16.4 \\
Implicit partial SVD  & 103.2 & 16.3 \\
Randomized SVD        & 10.9  & 15.4 \\
Stack + QR-SVD        & 15.0  & 22.6 \\
FLoRIST              & 64.6  & 25.2 \\
\hline
\multicolumn{3}{c}{\textit{CNN: 0.446M}} \\
RBLA                 & 1.2   & 11.1 \\
RBLA+ (QR-SVD)        & 11.6  & 21.4 \\
Dense full SVD        & 16.0  & 24.1 \\
Dense partial SVD     & 91.1  & 27.2 \\
Implicit partial SVD  & 119.4 & 27.0 \\
Randomized SVD        & 20.1  & 23.3 \\
Stack + QR-SVD        & 25.6  & 27.5 \\
FLoRIST              & 83.3  & 32.1 \\
\hline
\multicolumn{3}{c}{\textit{VGG9: 0.209M}} \\
RBLA                 & 1.3   & 4.9  \\
RBLA+ (QR-SVD)        & 3.5   & 11.2 \\
Dense full SVD        & 179.8 & 43.5 \\
Dense partial SVD     & 50.5  & 25.0 \\
Implicit partial SVD  & 42.8  & 13.4 \\
Randomized SVD        & 8.6   & 12.5 \\
Stack + QR-SVD        & 12.8  & 16.6 \\
FLoRIST              & 29.6  & 20.1 \\
\hline
\multicolumn{3}{c}{\textit{TinyBERT: 0.291M}} \\
RBLA                 & 2.0   & 8.9  \\
RBLA+ (QR-SVD)        & 4.6   & 16.9 \\
Dense full SVD        & 152.3 & 43.3 \\
Dense partial SVD     & 37.8  & 45.0 \\
Implicit partial SVD  & 13.6  & 20.0 \\
Randomized SVD        & 12.0  & 19.3 \\
Stack + QR-SVD        & 15.1  & 25.2 \\
FLoRIST              & 40.6  & 30.8 \\
\hline
\end{tabular*}
\vspace{-4mm}
\end{table}
\begin{table*}[t]
\centering
\caption{Analytical extraction costs for a fixed aggregated factor pair.}
\label{tab:aggregation_complexity}

\small
\setlength{\tabcolsep}{4pt}
\renewcommand{\arraystretch}{1.05}

\begin{tabular*}{\textwidth}{
    @{\extracolsep{\fill}}lll@{}
}
\toprule
Extraction
& Server computation
& Additional working storage \\
\midrule
RBLA (rank prefix)
& $\mathcal{O}(R)$
& $\mathcal{O}(R)$\\

RBLA+ (SP+)
& $\mathcal{O}((m+n)R^2+R^3)$
& $\mathcal{O}((m+n)R+R^2)$ \\

Dense product and full SVD
& $\mathcal{O}(mnR+mn\min\{m,n\})$
& $\mathcal{O}(mn+(m+n)\min\{m,n\})$ \\
\bottomrule
\end{tabular*}
\end{table*}

\subsubsection{Measured cost and accuracy tradeoffs}

Table~\ref{tab:accuracy-extraction-cost} compares the CPU adapter-extraction and output-preparation costs of RBLA+ with alternative decomposition methods. The benchmarks use synthetic adapter factors matching MLP, CNN, VGG9, and TinyBERT, comprising 3, 7, 9, and 17 factor pairs, respectively. Each configuration retains its prescribed client ranks, scaling coefficients, and actual matrix widths for convolutional layers. RBLA+ and the dense, implicit, and randomized SVD baselines process the same bounded-width aggregate. Stack + QR-SVD and FLoRIST instead process stacked factors constructed from the same synthetic client uploads. The measured phase includes decomposition, dense matrix construction where required, and preparation and retention of ten independent client-sized outputs. Measurements use FP32 on an AMD Ryzen 9 9950X processor, with PyTorch 2.3.1 and SciPy 1.16.0. We report the median of three timed repetitions after one warmup and the median of three separate memory measurements. The memory metric is the sampled peak process-RSS increment above the input-resident, using a nominal sampling interval of 1\,ms. Both partial-SVD implementations use ARPACK with tolerance $10^{-6}$ and initialization seed 41. Requests for the complete spectrum use full SVD in the dense baseline and QR-SVD in the implicit baseline. Randomized SVD uses an oversampling parameter of 10 and two power iterations. RBLA+ applies compact QR-SVD to the bounded-width factor pair, whereas Stack + QR-SVD applies the same decomposition to the wider stacked factors. FLoRIST retains its native decomposition and energy threshold of $0.9$. Across the four configurations, RBLA+ requires 3.5--11.6\,ms and has the lowest processing time and sampled $\Delta$RSS among the listed methods. Stack + QR-SVD requires 12.8--25.6\,ms, while FLoRIST requires 29.6--83.3\,ms. The MLP and CNN results also illustrate that iterative partial SVD need not outperform full SVD when many singular components are requested. In the CNN configuration, six of the seven layers request all $\min(m,n)$ singular components, while the remaining layer requests 192 components from a $384\times384$ matrix. These large target widths help explain why dense full SVD is faster than both partial-SVD implementations in this configuration. Despite the relatively low full-SVD cost in the MLP and CNN configurations, RBLA+ remains faster than all tested decomposition alternatives across the four configurations. Its compact QR/core-SVD introduces additional work over RBLA's direct prefix extraction, but the measured increase is limited to 2.2--10.4\,ms, with total extraction and output-preparation times of 3.5--11.6\,ms. RBLA+ also has the lowest sampled $\Delta$RSS among the compared decomposition methods. Thus, replacing direct prefixes with spectral extraction introduces a modest preparation overhead while avoiding the higher costs of the tested decomposition alternatives. This additional processing provides Frobenius-optimal rank-specific reconstruction without increasing client adapter sizes. At the prescribed client ranks, RBLA+ also outperforms RBLA in 24 of the 26 settings reported in Tables~\ref{tab:correct-rank-assignment} and~\ref{tab:mismatched-rank-assignment}, supporting the practical value of this additional server work. We also compare the cost of RBLA+ against stacking matrices. Exact stacking increases the internal width from the maximum client rank to the sum of client ranks, increasing both factorization work and memory requirements. Larger decomposition problems may also pose solver-dependent numerical challenges, particularly when the inputs are poorly scaled or ill-conditioned. Depending on the precision and decomposition backend, convergence difficulties can require additional iterations or fallback computations. We retain RBLA's compact aggregation to bound the internal representation by the maximum client rank and limit the associated preparation costs.

\subsubsection{Analytical extraction cost} Table~\ref{tab:aggregation_complexity} compares the work needed to prepare client adapters from the same $m\times R$ and $R\times n$ aggregated factors, where $m=d_{\mathrm{out}}$, $n=d_{\mathrm{in}}$, and $R\le\min(m,n)$. Prefix truncation require negligible preparation but hard to attain the optimal reconstruction. SP+ uses two thin QR factorizations and a core SVD to attain the minimum Frobenius reconstruction error, with a single decomposition serving all requested ranks; the full-SVD reference instead forms $BA$ explicitly and performs an economy-size SVD before truncation.

\subsubsection{Adapter communication payload}
The communication factors in Table~\ref{tab:correct-rank-assignment} and~\ref{tab:mismatched-rank-assignment} represent the analytical adapter-only downlink needed to serve the prescribed client ranks. Under the evaluated client-sized
transmission policy, RBLA, RBLA+, FlexLoRA, Zero Padding, Rep.~Padding, and FLoRIST transmit the same number of adapter entries when client ranks, target layers, and precision are matched. For a linear layer, client $i$ receives $(m+n)r_i$ scalars. With $K$ clients, the total payload is therefore $P_{\mathrm{compact}}=(m+n)\sum_{i=1}^{K}r_i$. Convolutional modules are counted using their actual factor dimensions. FLoRA avoids matrix decomposition but sends the full stacked factors to each recipient. With stacked width $S=\sum_{i=1}^{K}r_i$ and the same $K$ clients contributing and receiving updates, its total downlink is $P_{\mathrm{FLoRA}}=K(m+n)S=K P_{\mathrm{compact}}$. Thus, the ten-client experiments yield the reported $1\times$ and $10\times$ payload factors. This difference reflects the width of the representation delivered to each client. RBLA+ performs spectral extraction at the server and transmits only the factors matching each client's budget. Spectral reordering and client-scale conversion preserve these prescribed output dimensions. Its reconstruction improvement therefore entails no additional adapter payload over direct RBLA extraction at fixed client ranks. The payload comparison counts the total entries sent to individual recipients and excludes the initial pretrained-model distribution, serialization metadata, and network-protocol overhead.

\subsubsection{Interpretation and scope}

The reconstruction experiments confirm SP+'s fixed-aggregate optimality guarantee within numerical precision. The cross-layer analysis complements this finding by examining raw prefixes before spectral resetting. At the inspected checkpoints, the RBLA+ training path yields lower median excess reconstruction errors, with each prefix measured against the optimum for its own reference aggregate. Accuracy improves over RBLA in both low-rank training from scratch and pretrained-model adaptation. The cross-broadcast results further show that extraction's effect on client side predictive performance depends on the training path: spectral broadcast improves client accuracy on MNLI along the RBLA+ path but reduces it along the RBLA path. Reconstruction optimality thus does not guarantee an immediate accuracy gain for every aggregated LoRA adapter. CPU benchmarks and communication comparisons show that RBLA+ delivers these empirical gains with modest additional preparation costs over RBLA and unchanged client-sized payloads, showing its accuracy--cost tradeoff under prescribed rank budgets.

\section{Conclusion}
We show that directly truncating LoRA factors in RBLA can introduce avoidable reconstruction error. RBLA+ addresses this limitation using an established compact QR/core-SVD construction for Frobenius-optimal rank-specific extraction and bidirectional LoRA scaling alignment. Experiments show that compact extraction matches the optimal reconstruction reference within numerical precision. Evaluation results revealed that compared to state-of-the-art methods, RBLA+ can improve accuracy by up to 3.30\%, and by up to 40.79 percentage points over RBLA on CIFAR-10, with a 90\% smaller downlink payload than FLoRA in the ten-client setting. Across MLP, CNN, VGG9, and TinyBERT configurations, RBLA+ adds 2.2--10.4 ms of server processing over RBLA, with total times of 3.5--11.6 ms. These gains preserve the prescribed client ranks and adapter sizes.
\bibliographystyle{IEEEtran}
\bibliography{reference}

\end{document}